\documentclass[runningheads]{llncs}
\usepackage{graphicx}

\usepackage{tikz}
\usepackage{comment}
\usepackage{amsmath,amssymb} 
\usepackage{color}

\usepackage[accsupp]{axessibility}  

\usepackage[width=122mm,left=12mm,paperwidth=146mm,height=193mm,top=12mm,paperheight=217mm]{geometry}

\usepackage{epsfig}
\usepackage{graphicx}
\usepackage{amsmath}
\usepackage{amssymb}
\usepackage{upgreek}
\usepackage{booktabs}
\usepackage{diagbox}
\usepackage{multirow}
\usepackage{url}

\usepackage{pifont}
\newcommand{\cmark}{\ding{51}}%
\usepackage{colortbl}

\definecolor{lightgray}{rgb}{0.9, 0.9, 0.9}

\newlength\savewidth\newcommand\shline{\noalign{\global\savewidth\arrayrulewidth
  \global\arrayrulewidth 1pt}\hline\noalign{\global\arrayrulewidth\savewidth}}
 
\usepackage{capt-of}
\usepackage[colorlinks]{hyperref}

\begin{document}
\pagestyle{headings}
\mainmatter

\title{Conditional DETR V2: 
Efficient Detection Transformer
with Box Queries} 

\selectfont

\titlerunning{Conditional DETR V2}
%
\author{Xiaokang Chen\inst{1} \and
Fangyun Wei\inst{2} \and
Gang Zeng\inst{1} \and
Jingdong Wang\inst{3}}
\authorrunning{Chen et al.}
%
\institute{\textsuperscript{\rm 1} Key Laboratory of Perception (MoE), School of AI, Peking University \\ \textsuperscript{\rm 2} Microsoft Research Asia \  \textsuperscript{\rm 3} Baidu}
\maketitle

\begin{abstract}
In this paper,
we are interested in Detection Transformer (DETR),
an end-to-end object detection approach
based on a transformer encoder-decoder architecture
without hand-crafted postprocessing, such as NMS.
Inspired by Conditional DETR,
an improved DETR with fast training convergence,
that presented box queries (originally called spatial queries)
for internal decoder layers,
we reformulate the object query
into the format of the box query
that is a composition
of the embeddings
of the reference point
and the transformation
of the box with respect
to the reference point.
This reformulation 
indicates the connection
between the object query in DETR
and the anchor box that is widely studied
in Faster R-CNN.
Furthermore,
we learn the box queries from the image content,
further improving the detection quality
of Conditional DETR
still with fast training convergence.
In addition,
we adopt the idea of axial self-attention to save the memory cost and accelerate the encoder.
The resulting detector, called Conditional DETR V2,
achieves better results
than Conditional DETR,
saves the memory cost
and runs more efficiently.
For example,
for the DC$5$-ResNet-$50$ backbone,
our approach
achieves $44.8$ AP with $16.4$ FPS on the COCO $val$ set and compared to Conditional DETR, it
runs $1.6\times$ faster, saves $74$\% of the overall memory cost, and improves $1.0$ AP score.

\end{abstract}

\section{Introduction}
Object detection is a problem of localizing the objects 
and 
predicting their categories.
It is a fundamental and challenging problem in computer vision~\cite{chen2022context,tang2022not,tang2022point,chen2021semi,chen2020bi,chen20203d}
and has many practical applications, such as robot navigation and autonomous driving.
Deep learning has been the dominant solution to object detection.
The early framework regresses the object box by starting from a default box (or anchor box)
and optionally iteratively refines the box regression,
e.g., Faster R-CNN~\cite{ren2015faster}.
Recently, detection transformer (DETR)
exploits the transformer decoder for object detection,
with introducing the object queries (learnable positional embeddings).
It is an end-to-end solution
without the post-processing, such as non-maximum suppression (NMS).

DETR typically needs $500$ epochs to achieve satisfactory performance.
Several follow-up works,~\cite{SunCYK20,GaoZWDL21,wang2021anchor,meng2021conditional,DaiCLC20,ZhuSLLWD20} develop DETR variants
to accelerate the training process.
Conditional DETR~\cite{meng2021conditional} and Anchor DETR~\cite{wang2021anchor} identify that the decoder cross-attention 
has limited capability to exploit the spatial attention
and designs the new query format
to boost the capability.

\begin{figure}[ht]
    \centering
    \includegraphics[width=1.0\textwidth]{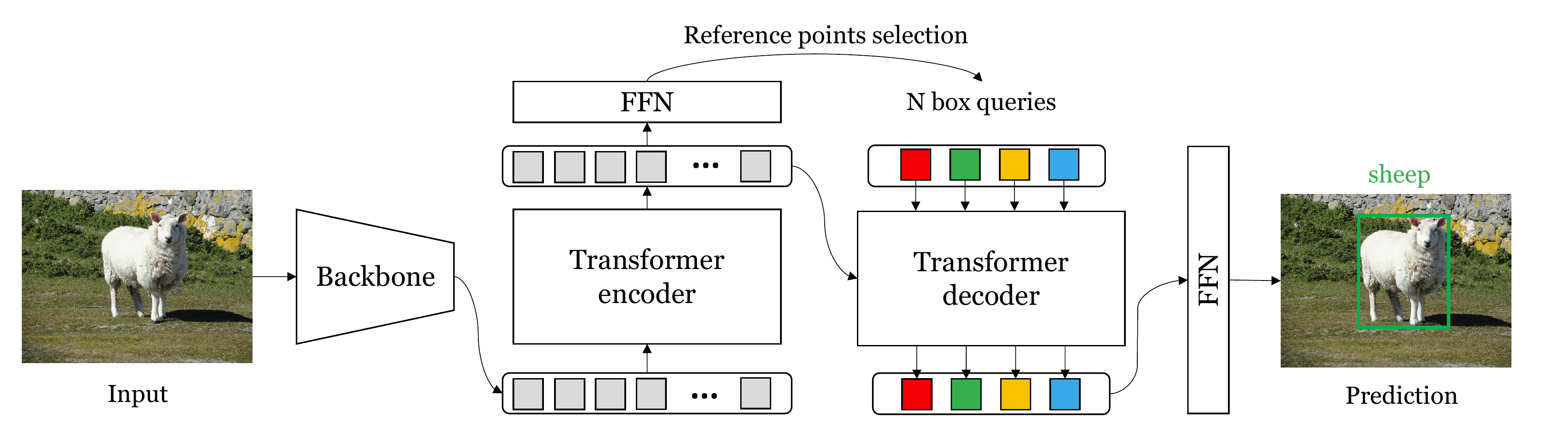}
    \captionof{figure}{The overall architecture. 
(1) We use a conventional CNN backbone ($e.g.$, ResNet-$50$) to extract the feature of an input image. (2) The learned embedding is fed into the transformer encoder layers to model the global dependencies between the inputs. (3) Then, we predict candidate boxes from the encoder embedding and select reference points according to the classification score of these boxes. The input of the decoder is the concatenation of the box query and the content query. The box query consists of the embedding of the selected reference points and the transformation predicted from the corresponding encoder embedding of these reference points. The content query is initialized from the selected candidate boxes. 
(4) We pass each output embedding of the
decoder to a shared feed-forward network (FFN) that predicts either a detection (class and bounding box) or a "no object" class.
    }
    \label{fig:overall-arch}
\vspace{-.2cm}
\end{figure}

Inspired by the spatial queries (we call box queries)
that Conditional DETR introduces for internal decoder layers,
we propose to study the object query (positional encodings),
the input of the first decoder layer,
and reformulate it as the form
of the box query
in the embedding space.
The box query
is a composition
of the embeddings of the reference point 
and the transformation
of the box with respect
to the reference point.
The box query form in some sense
builds the connection between
DETR and Faster R-CNN
that both use the boxes as one input.
There are several differences.
The box query is in the embedding space
and the anchor box is in a form of $2$D coordinates.
The box query is explored through the attention mechanism
to find the extremities
for box regression
and the regions inside the object 
for classification,
while the anchor box is used as the initial guess.

We predict the box queries
from the image content
for each image,
instead of 
learning them as model parameters.

Specifically, we predict the probability of belonging to an object for the points in the encoder embedding, and select the top-scored points as the reference points. 
The transformation that contains the scale information of the object is predicted from the corresponding encoder embedding. The image-dependent box query helps locate the object and improve the performance.

In addition,
we adopt the idea of axial self-attention~\cite{HoKWS19} and propose the Horizontal-Vertical Attention. For each query embedding, we perform attention in the horizontal and vertical directions in parallel. Experiments show that the Horizontal-Vertical Attention could save the memory cost and accelerate the encoder, especially for high-resolution backbones, such as DC$5$-ResNet-$50$ and DC$5$-ResNet-$101$.

\section{Related Work}
\vspace{0.1cm}
\noindent\textbf{Anchor-based object detectors.} 
Anchor plays an important role in object detection. Faster R-CNN~\cite{ren2015faster} first introduces the anchor boxes to make dense predictions. A series of works such as SSD~\cite{LiuAESRFB16}, RetinaNet~\cite{LinGGHD20}, YOLO~\cite{RedmonF17,JA18,BWL20} and Cascade R-CNN~\cite{CaiV18} are proposed to make object detection more efficient and accurate. To avoid hand-crafted anchor design, some recent methods, e.g. FCOS~\cite{TianSCH19}, CenterNet~\cite{ZWP19}, CornerNet~\cite{law2018cornernet} and ExtremeNet~\cite{zhou2019bottom}, use anchor points to represent objects, these methods are also known as anchor-free detectors.

In this paper, we reformulate the object query in DETR into the format of the box query that consists of the embeddings of the reference point and the transformation of the box with respect to the reference point. This reformulation indicates the connection between the object query in DETR and 
 the anchor box in Faster R-CNN.



\vspace{0.1cm}
\noindent\textbf{DETR and its variants.} 
DETR successfully applies transformers
to object detection and removes the need for non-maximum suppression. The object queries (or called position embeddings) in DETR are responsible for the category prediction and box regression of each object. Some works~\cite{ZhuSLLWD20,wang2021pnp} aim to solve the high computation complexity issue caused by the global encoder self-attention in DETR. Deformable DETR~\cite{ZhuSLLWD20} designs the 
sparse attention that each query only attends to keys in a local region. PNP-DETR~\cite{wang2021pnp} proposes the poll-and-pool strategy to reduce the spatial redundancy in the input, resulting in a smaller input size for the encoder.

DETR also suffers from another issue, the slow training convergence, which attracts a lot of attention.
TSP~\cite{SunCYK20} eliminates the cross-attention modules
and combines the FCOS and R-CNN-like detection heads.
Deformable DETR~\cite{ZhuSLLWD20}
proposes deformable attention and uses multiple level features to accelerate the converge. 
SMCA\cite{GaoZWDL21} modulates the DETR multi-head global cross-attentions with gaussian maps around a few (shifted) centers to focus more on the restricted local regions.

Anchor DETR~\cite{wang2021anchor} argues that object query in DETR could not focus on a specific region and replace object queries with anchor points. Conditional DETR~\cite{meng2021conditional} proposes to search the extremity and distinct regions of the object through learning the conditional spatial queries from the decoder embeddings. However, the spatial query of the first layer could not find such regions well due to the lack of the image content in the query. To solve this issue, we propose the box query and initialize it from the image content, which could find extremity regions since the first layer.

\vspace{0.1cm}
\noindent\textbf{Efficient attention mechanism.} The global self-attention~\cite{VaswaniSPUJGKP17} could model the global dependencies between the inputs, while the computation complexity is quadratic to the input size. Many efficient attention mechanisms have been designed in the NLP field to improve  the memory and computation efficiency of self-attention, such as
Sparse Transformer~\cite{CGRS19},
Linformer~\cite{WLKFM20},
Performer~\cite{Krzysztof20},
RFA~\cite{PPYSK21},
Nystr{\"{o}}mformer~\cite{XiongZCTFLS21},
Reformer~\cite{KKL20},
Routing Transformer~\cite{RSVG21},
BigBird~\cite{ZGDAAOPRWYA20},
Longformer~\cite{BPA20},
Transformer-XL~\cite{DaiYYCLS19}. In the vision field, some methods~\cite{ramachandran2019stand,vaswani2021scaling} restrict the attention region
of the key for each query instead of the whole region. 
For example, Swin Transformer~\cite{liu2021swin} and HRFormer~\cite{yuan2021hrformer} adopt the local self-attention mechanism and add interaction across different local windows. 
Axial self-attention~\cite{HoKWS19} applies the local window along the horizontal or vertical axis sequentially to achieve global attention. 
Following axial self-attention, we propose the Horizontal-Vertical Attention, where each query pixel attends to pixels that belong to the same row or column as it in parallel.

\section{Approach}
\subsection{Architecture}
We follow detection transformer (DETR)
to use the transformer encoder and decoder structure for object detection.
The architecture consists of
a CNN backbone,
a transformer encoder,
a transformer decoder,
and predictors for classification and box regression.
The transformer encoder aims to improve the content embeddings output from the CNN backbone.
It is a stack
of multiple encoder layers,
where each layer mainly consists of 
a self-attention layer
and a feed-forward layer.

The transformer decoder
is a stack of decoder layers.
Each decoder layer
is composed of three main layers:
(1) a self-attention layer
with the embeddings,
outputted from the previous decoder layer as the input,
(2) a cross-attention layer,
with the output of the self-attention layer
as the queries
and the output of the encoder
forming the keys and the values,
and (3) a feed-forward layer
predicting the box and the category score.

A candidate box 
is predicted
from each decoder embedding as follows,
\begin{align}
    \hat{\mathbf{b}} = \operatorname{sigmoid}(\operatorname{FFN}(\mathbf{f}) + [{\mathbf{s}}^\top~0~0]^\top)
    \label{eqn:boxprediction}.
\end{align}
Here, $\mathbf{f}$ is the decoder embedding.
$\hat{\mathbf{b}}$ is a four-dimensional vector
$[b_{cx}~b_{cy}~b_{w}~b_{h}]^\top$,
consisting of
the box center,
the box width
and the box height.
${\mathbf{s}}$
is the unnormalized $2$D coordinate of 
the reference point,
and is $(0,0)$
in the original DETR. 
The classification score for each candidate box
is also predicted from
the decoder embedding
through another FNN,
$\mathbf{e} = \operatorname{FFN}(\mathbf{f})
$.

\subsection{Box Query}
\noindent\textbf{Object query.}
The original DETR learns~\emph{object queries},
\begin{align}
    \{\mathbf{o}_1, \dots, \mathbf{o}_N\},
\end{align}
where each query is a high-dimensional embedding vector.
These object queries are 
model parameters
and are the same for all the images.
In DETR~\cite{CarionMSUKZ20},
they are also called positional encodings,
and 
initially used to encode the information
of the positions at which the objects
might appear.
The object queries are combined into the decoder embeddings,
forming the queries of the self-attention and cross-attention layers in the decoder.

\vspace{0.1cm}
\noindent\textbf{Box query.}
Conditional DETR 
learns spatial queries,
which we call box queries in this paper,
from the decoder output embeddings
and use them to
replace the object queries 
for the decoders except for the first decoder layer.
The box queries are combined
with the content queries
through concatenation 
instead of addition in the original DETR.
Accordingly,
the keys are formed
by concatenating the positional embeddings
and the encoder output embeddings.

The box queries $\mathbf{p}_q$ 
are predicted by transforming the 2D reference point
in the embedding space
with the transformation of the box with respect to the
reference point: 
\begin{align}
    \mathbf{p}_q =   \boldsymbol{\uplambda}_q \odot {\mathbf{p}}_s.\label{eqn:conditionalspatialembedding}
\end{align}
Here,
$\mathbf{p}_s$ is the projected positional embedding from the reference point and  $\boldsymbol{\uplambda}_q$
is a transformation in the embedding space.
$\odot$
is the element-wise multiplication operator.

\vspace{0.1cm}
\noindent\textbf{Learning initial box queries as model parameters.}
We use the box queries
to replace the object queries
that Conditional DETR still uses in the first decoder layer.
We learn $2$D references points,
$\{(c_{x1}, c_{y1}), \dots,
(c_{xK}, c_{yK})$\},
and use the sinusoidal positional embedding to form the corresponding embeddings $\{\mathbf{p}_{s1}, \dots, \mathbf{p}_{sK}\}$. $\boldsymbol{\uplambda}_q$ is set as the model parameter that is randomly initialized.

\begin{figure}[t]
\centering
\includegraphics[width=.6\textwidth]{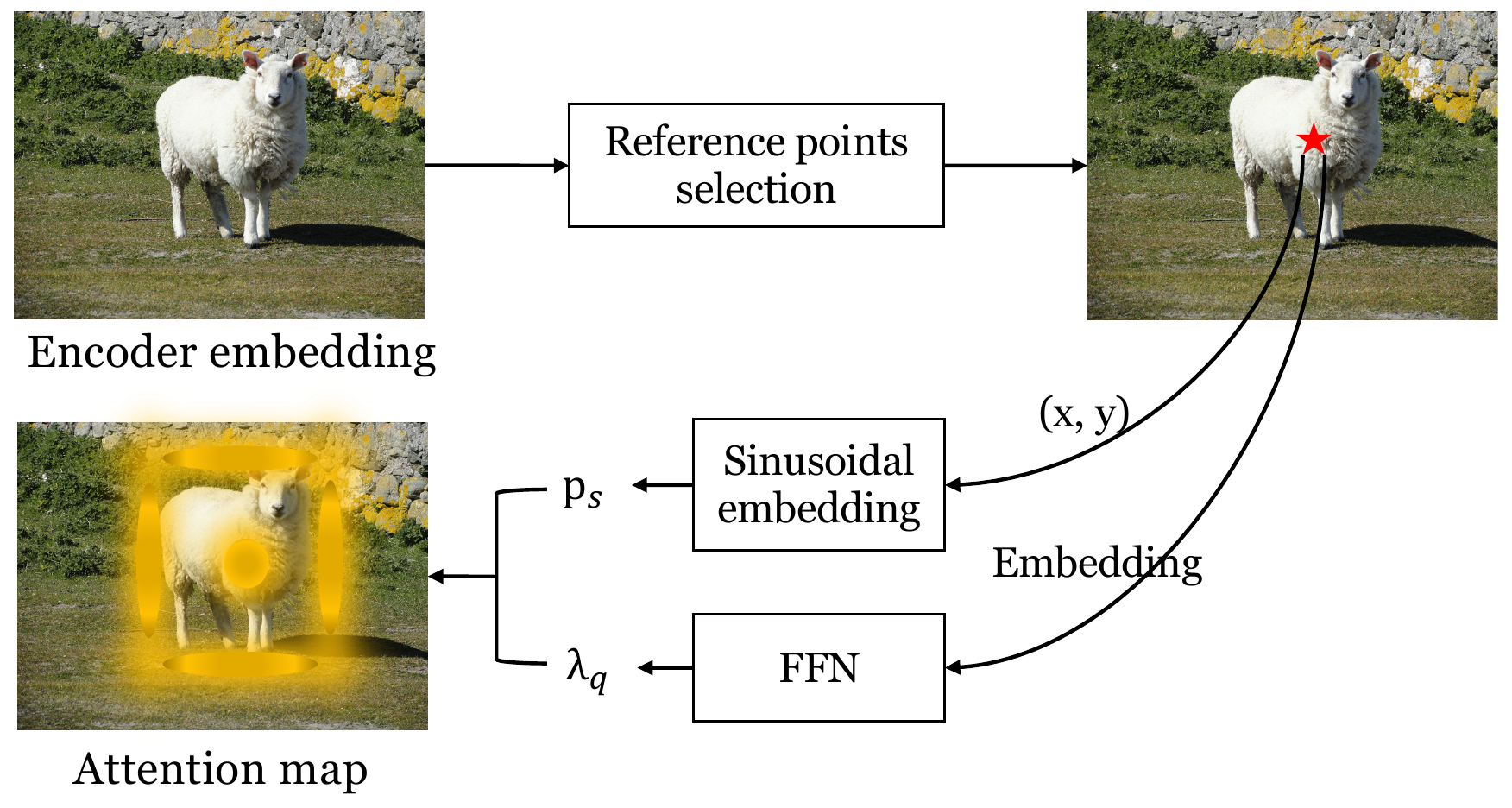}
 
   \caption{Pipeline of box query initialization. The box query is a composition of the embeddings of the box center (reference point) and the box size. The attention map is a toy example, not the actual output from the model.}
\label{fig:box-query-arch}
\end{figure}

\vspace{0.1cm}
\noindent\textbf{Predicting initial box queries from image content.}
The pipeline is shown in Figure~\ref{fig:box-query-arch}.
We propose to 
select several positions
as reference points,
$\{(c_{x1}, c_{y1}), \\ \dots, 
(c_{xK}, c_{yK})$\}
from the image content. 
We then build the reference embeddings from these points
and the transformation
for each reference point
according to the corresponding encoder embedding $\mathbf{\hat X}$
as the following,
\begin{align}
\boldsymbol{\uplambda}_q = \operatorname{FFN}(\mathbf{\hat x}{(c_{x}, c_{y})}),
\end{align}
where $\mathbf{\hat x}{(c_{x}, c_{y})}$ is the encoder embedding at the position $(c_{x}, c_{y})$. $(c_{x}, c_{y})$ is the normalized position which ranges from $0$ to $1$. We select the reference positions
by classifying each position.
The classification score 
is also predicted from
the encoder embedding
through an FFN,
\begin{align}\mathbf{s} = \operatorname{FFN}(\mathbf{\hat x}{(c_{x}, c_{y})}).
\end{align}
We perform binary classification here and $\mathbf{s}$ is a $2$-d vector, indicating the possibility of being an object ($1$) or non-object ($0$). We rank all the positions 
according to their 
probability of being an object
and the top-scored predictions are selected.

The purpose of using image content is different from the two-stage strategy in Deformable DETR. They generate the initial positions and box sizes from the image content, then sample the key set around those positions. The final box size is predicted based on the initial box size, which is a form of residual prediction and easier to optimize.
By contrast, we project the position and box information to high-dimensional space and use them to form the box query. The box query could scale the range that we search for the object according to its size.

\vspace{.1cm}
\noindent\textbf{Initial content queries from image.}
In the internal decoder layers,
the content query,
output from the previous decoder layer,
is used to predict the box,
indicating that
the content query also contains the box information.
In light of this,
we initialize the content query for the first decoder layer
by incorporating the box information.

We consider two ways.
One is to initialize the content query 
from the encoder embeddings:
\begin{align}
    \mathbf{c}_q = \operatorname{FFN}(\mathbf{\hat x}{(c_{x}, c_{y})}). 
\end{align}
The other one uses the box estimated from the position:
\begin{align}
\label{eq:content-query-init}
    \mathbf{c}_q = \operatorname{FFN}(\operatorname{PE}(\mathbf{b}(c_{x}, c_{y})).
\end{align}
Here, $\mathbf{b}(c_{x}, c_{y})$
is the box estimated from the position $(c_{x}, c_{y})$:
$ \operatorname{sigmoid}(\operatorname{FFN}( \\ \hat{\mathbf{x}}(c_x, c_y)) + \sigma^{-1}([c_x~c_y~c_w~c_h]^\top))$,
and $\operatorname{PE}()$
is the positional embedding. $\sigma^{-1}$ means the inverse sigmoid function. $c_w$ and $c_h$ are hyper-parameters and we will study them in the experiment section.
Our results empirically indicate
that the two ways perform similarly and the latter way performs slightly better.

\subsection{Horizontal-Vertical Attention}
The encoder attention is  
formulated as follows, 
\begin{align}
    \operatorname{Atten}(\mathbf{x}_{uv}, \mathbf{X}, \mathbf{X})
    &= \sum_{i=1}^H\sum_{j=1}^W \mathbf{x}_{ij} 
    \alpha(\mathbf{x}_{uv}, \mathbf{x}_{ij}). \\
    \alpha(\mathbf{x}_{uv}, \mathbf{x}_{ij}) &= \frac{e^{\frac{1}{\sqrt{d}}\mathbf{x}_{uv}^\top \mathbf{x}_{ij}}}{\sum_{i=1}^H\sum_{j=1}^W e^{\frac{1}{\sqrt{d}}\mathbf{x}_{uv}^\top \mathbf{x}_{ij}}.}
\end{align}
Here $\mathbf{X}$ is the embedding output from the backbone, $e.g.,$ ResNet-$50$. We use one query $\mathbf{x}_{uv}$
at the position $(u, v)$
as an example. $d$ is the feature dimension.
The memory and computation complexities
for all the $HW$ queries are
quadratic 
with respect to the map size $HW$.

\begin{figure}[ht]
\centering
\includegraphics[width=.66\textwidth]{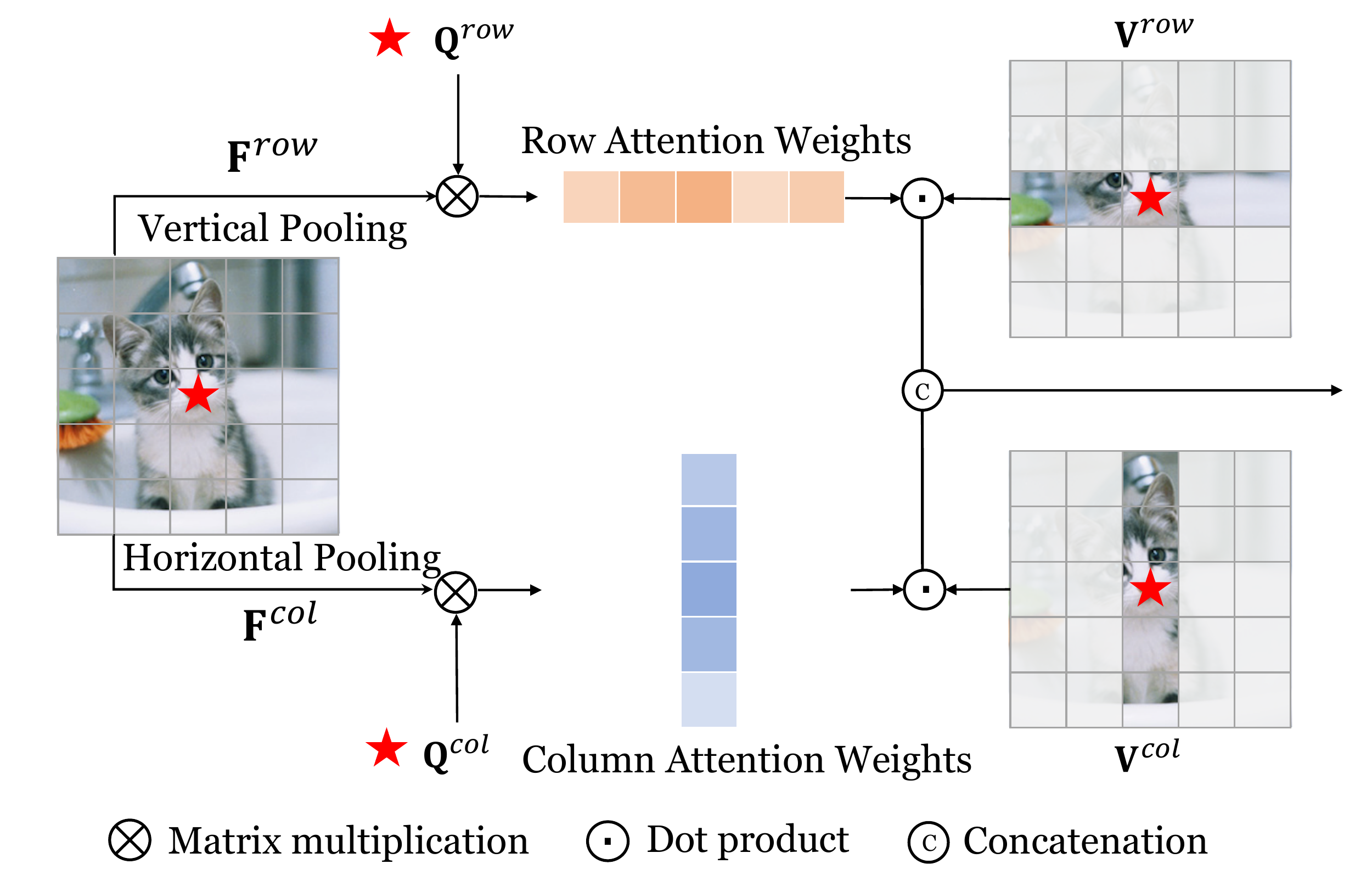}
 
   \caption{The proposed Horizontal-Vertical Attention. Each pixel will aggregate embeddings of the pixels in the same row or column. $\mathbf{Q}^{row}$ ($\mathbf{Q}^{col}$) is generated by the addition of the query embedding and the $1$D row (col) position encoding. $\mathbf{F}^{row}$ ($\mathbf{F}^{col}$) is generated by the addition of the pooled row (col) embedding and the $1$D row (col) position encoding. $\mathbf{V}^{row}$ ($\mathbf{V}^{col}$) is from the pixels that lie in the same row (col) as the query pixel.}
\label{fig:hvattn}
\end{figure}

We propose to decrease the number of keys
for complexity reduction, as shown in Figure~\ref{fig:hvattn}.
The idea is inspired by CCNet~\cite{huang2019ccnet} and axial self-attention~\cite{HoKWS19}.
We pool the embedding $\mathbf{X}$
along the vertical and
horizontal  directions respectively,
$\bar{\mathbf{x}}_{: j}
=\frac{1}{H}\sum_{i=1}^H \mathbf{x}_{i j}$,
and $\bar{\mathbf{x}}_{i :}
=\frac{1}{W}\sum_{j=1}^W \mathbf{x}_{i j}$,
and form $(H+W)$ keys:
\begin{align}
    \mathbf{K} = [\bar{\mathbf{x}}_{1 : }~\dots~\bar{\mathbf{x}}_{H :}
    ~\bar{\mathbf{x}}_{: 1}~\dots~\bar{\mathbf{x}}_{: W}].
\end{align}

The HV-attention process is as follows,
\begin{align}
    \operatorname{HV-Atten}(\mathbf{x}_{uv}, \mathbf{K}, \mathbf{X})
    = ~&
    \operatorname{Concat}(\sum_{i=1}^H
    \mathbf{x}_{iv} 
    \alpha(\mathbf{x}_{uv}, \bar{\mathbf{x}}_{i:}),  \nonumber \\
    & 
    \sum_{j=1}^W \mathbf{x}_{uj} 
    \alpha(\mathbf{x}_{uv}, \bar{\mathbf{x}}_{:j})).
\end{align}

\vspace{0.1cm}
\noindent \textbf{Complexity analysis.}
The computation complexity of HV-attention 
is reduced from quadratic to linear
with respect to $HW$.
The computational complexity of global self-attention and our Horizontal-Vertical Attention based on the feature map $\mathbf{F} \in \mathbb{R}^{d \times H \times W}$ are:

\begin{align}
\label{self-att-complex}
    &\Omega(\operatorname{Global-Atten}) = 4 HWd^2 + 2(HW)^2d, \\
    \label{rc-att-complex}
    &\Omega(\operatorname{HV-Atten}) = 7 HWd^2 + 2HWd(H+W),
\end{align}
where the coefficient of the first item increases from $4$ to $7$ because we apply two linear projections on the query, key and value. Eq. (\ref{self-att-complex}) is quadratic to $HW$ and is unaffordable when $HW$ is large.

\vspace{0.1cm}
\noindent \textbf{Comparison with CC Attention~\cite{huang2019ccnet}.}
The process of CC Attention is as follows,
\begin{align}
\mathbf{K}_{uv} = [\mathbf{x}_{1v},~\dots~\mathbf{x}_{Hv}, & \mathbf{x}_{u1}, ~\dots~\mathbf{x}_{uW}]. \\
    \operatorname{CC-Atten}(\mathbf{x}_{uv}, \mathbf{K}_{uv}, \mathbf{X})
    = ~&
    \sum_{i=1}^H
    \mathbf{x}_{iv} 
    \alpha(\mathbf{x}_{uv}, \mathbf{x}_{iv}) +   \nonumber \\
    & 
    \sum_{j=1}^W \mathbf{x}_{uj} 
    \alpha(\mathbf{x}_{uv}, \mathbf{x}_{uj}).
\end{align}

Here $\mathbf{X}$ is the embedding output from the backbone, $e.g.,$ ResNet-$50$. We use one query $\mathbf{x}_{uv}$
at the position $(u, v)$
as an example.  $\mathbf{K}_{uv}$ is the key set for the query position $(u, v)$.
There are two differences between our method and CC Attention. (1) We use a shared key set for all the queries, further saving the memory cost of the attention module.
For example,
$13$\% memory cost is saved
for the $256 \times 100 \times 150$ ($d \times H \times W$) input.  (2) We use concatenation instead of addition when aggregating the horizontal and vertical features, which brings $0.5$ AP improvement ($44.3 \rightarrow 44.8$) on the DC$5$-ResNet-$50$ backbone.

\vspace{0.1cm}
\noindent \textbf{Comparison with RCDA~\cite{wang2021anchor}.}
We also compare our Horizontal-Vertical Attention with the Row-Column Decoupled Attention (RCDA). RCDA decouples the $2$D key feature to
the row feature and the column feature by $1$D global average pooling, and performs the row attention and the column attention successively. Our proposed method is different from them in that: (1) We conduct row attention and column attention in a parallel way. (2) The value set we use is the feature of pixels that belong to the same row or column as the query pixel, while RCDA views all the pixels as the value set, bringing more computational cost.

\begin{figure*}[t]
\centering
\includegraphics[ width=0.19\linewidth]{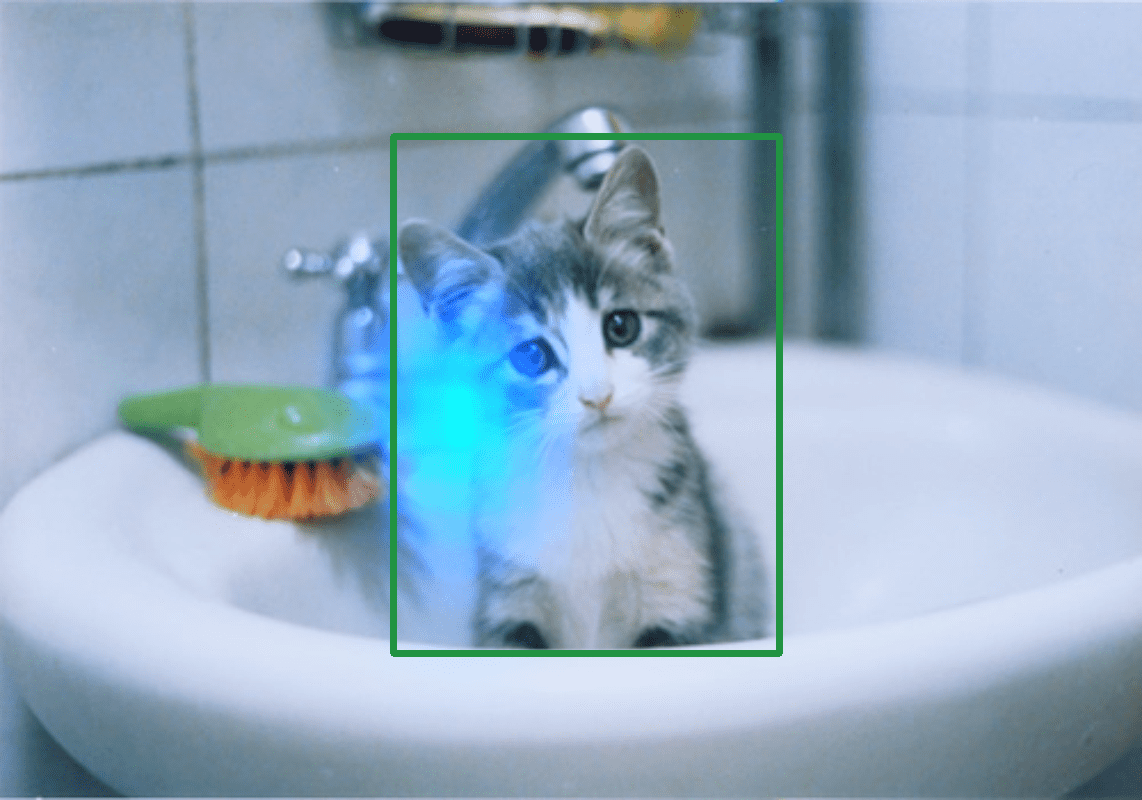}
\includegraphics[ width=0.19\linewidth]{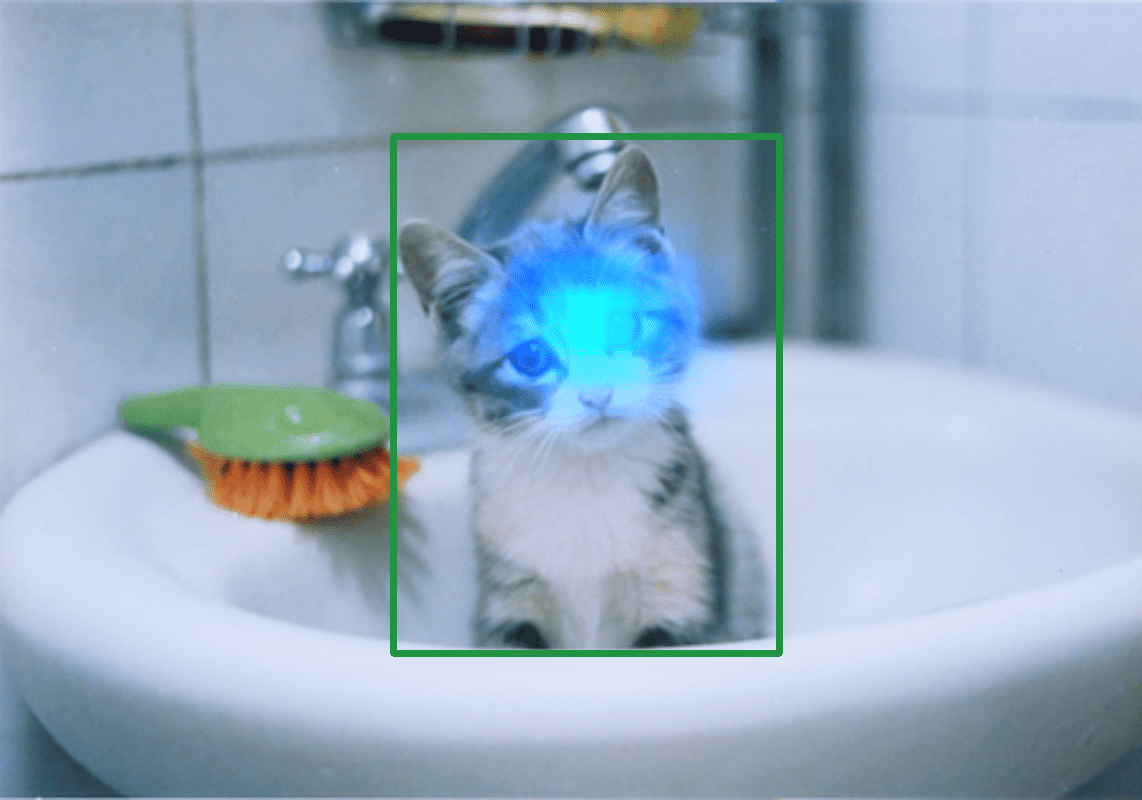}
\includegraphics[ width=0.19\linewidth]{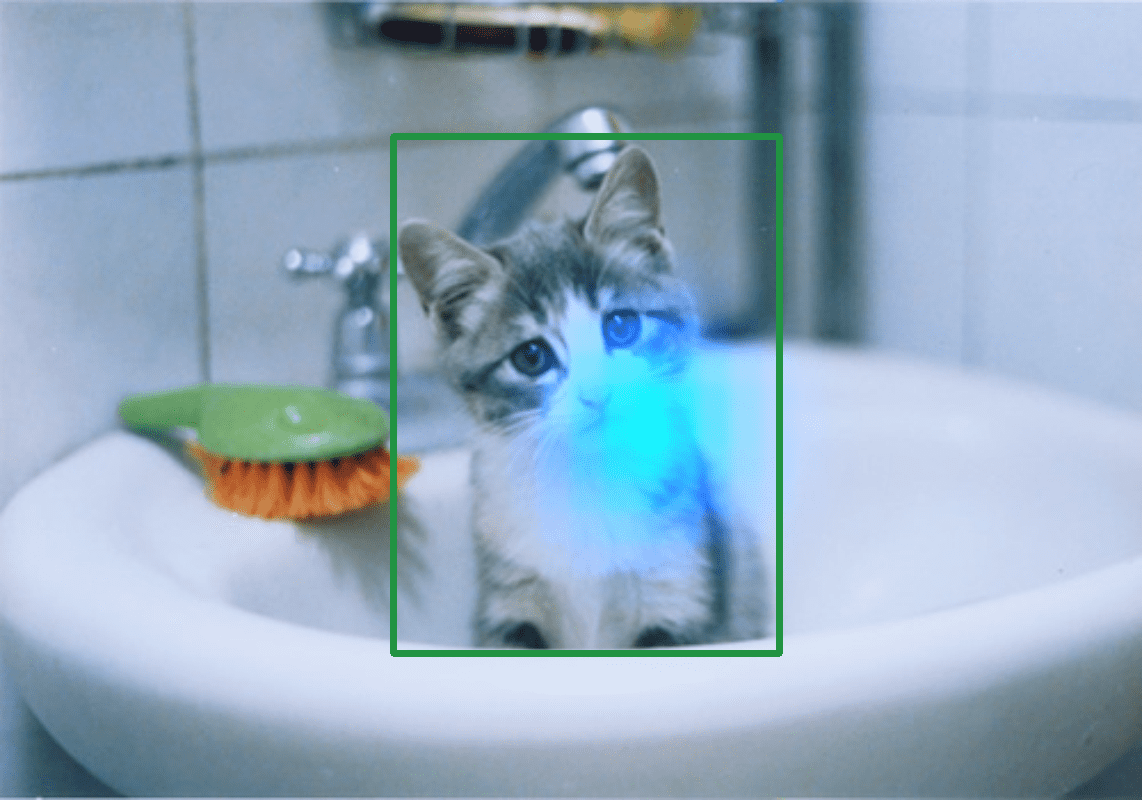}
\includegraphics[ width=0.19\linewidth]{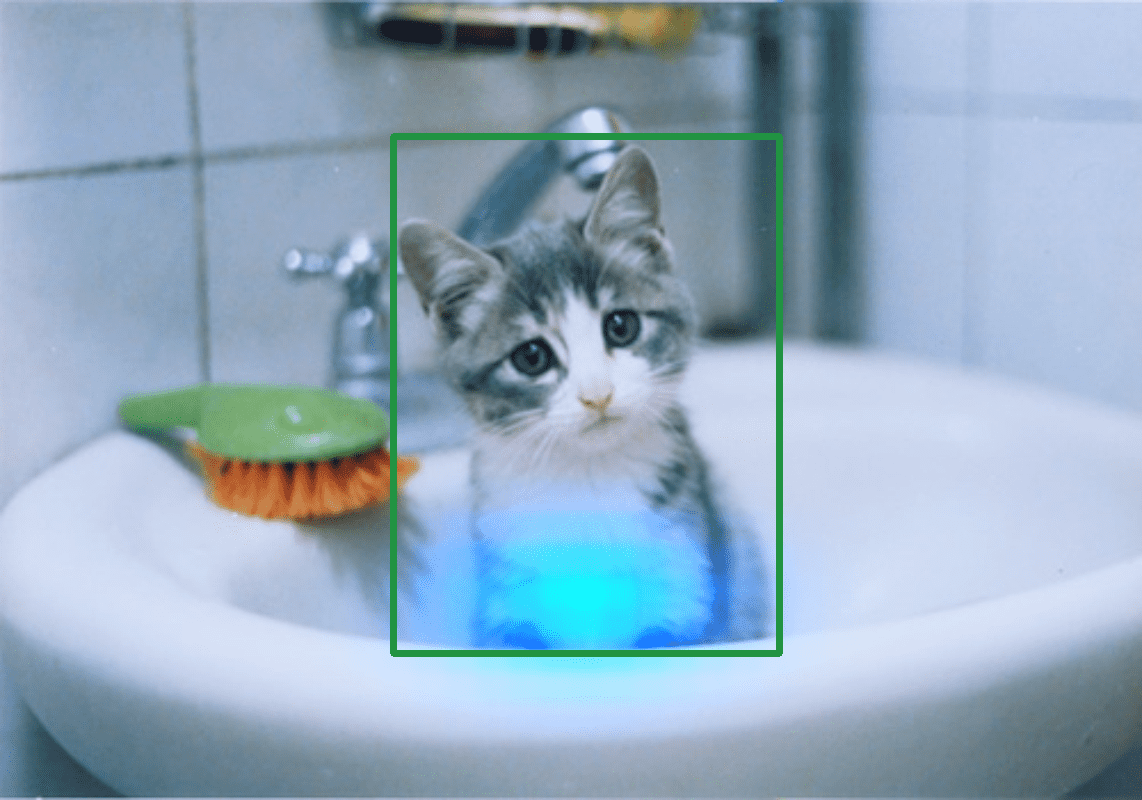}
\includegraphics[ width=0.19\linewidth]{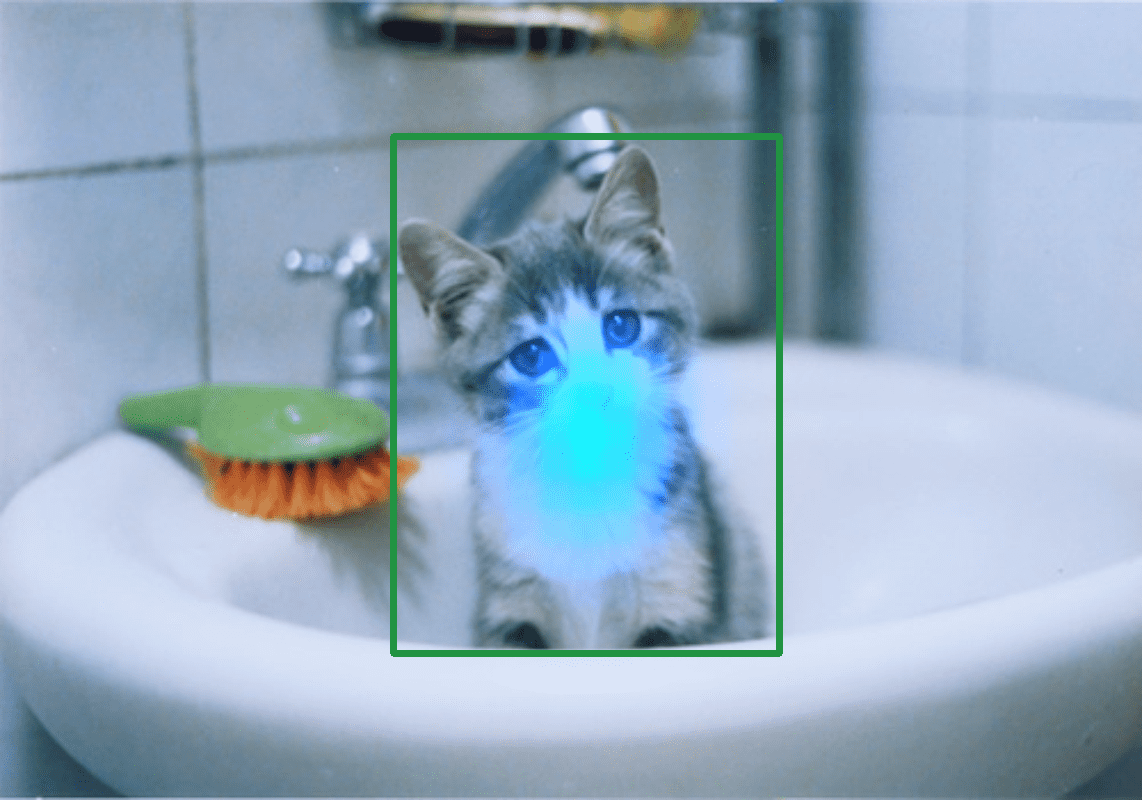}
\\
\includegraphics[ width=0.19\linewidth]{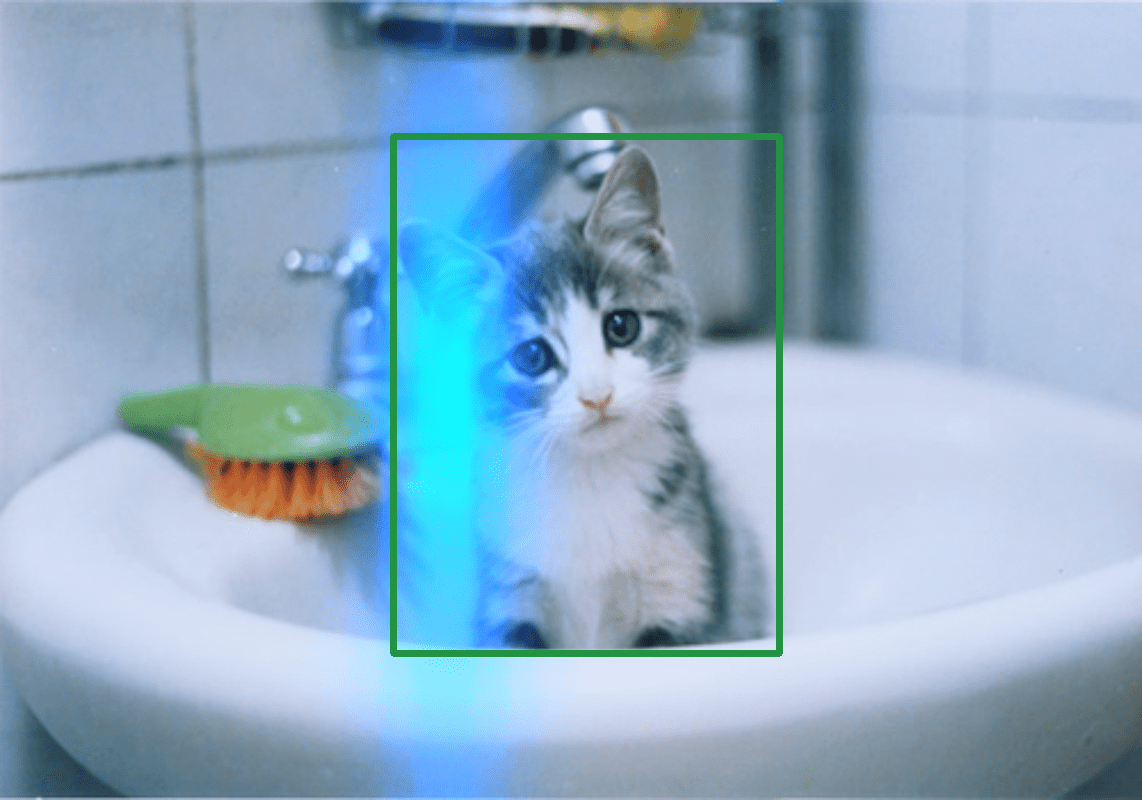}
\includegraphics[ width=0.19\linewidth]{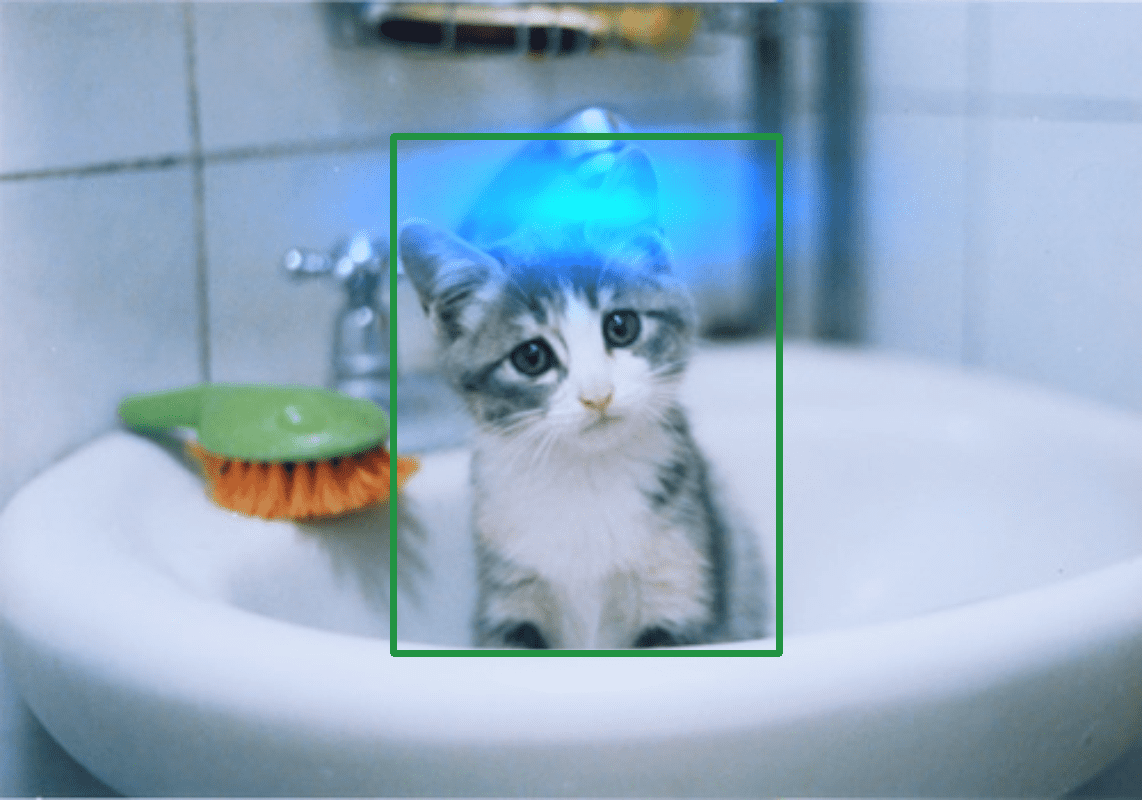}
\includegraphics[ width=0.19\linewidth]{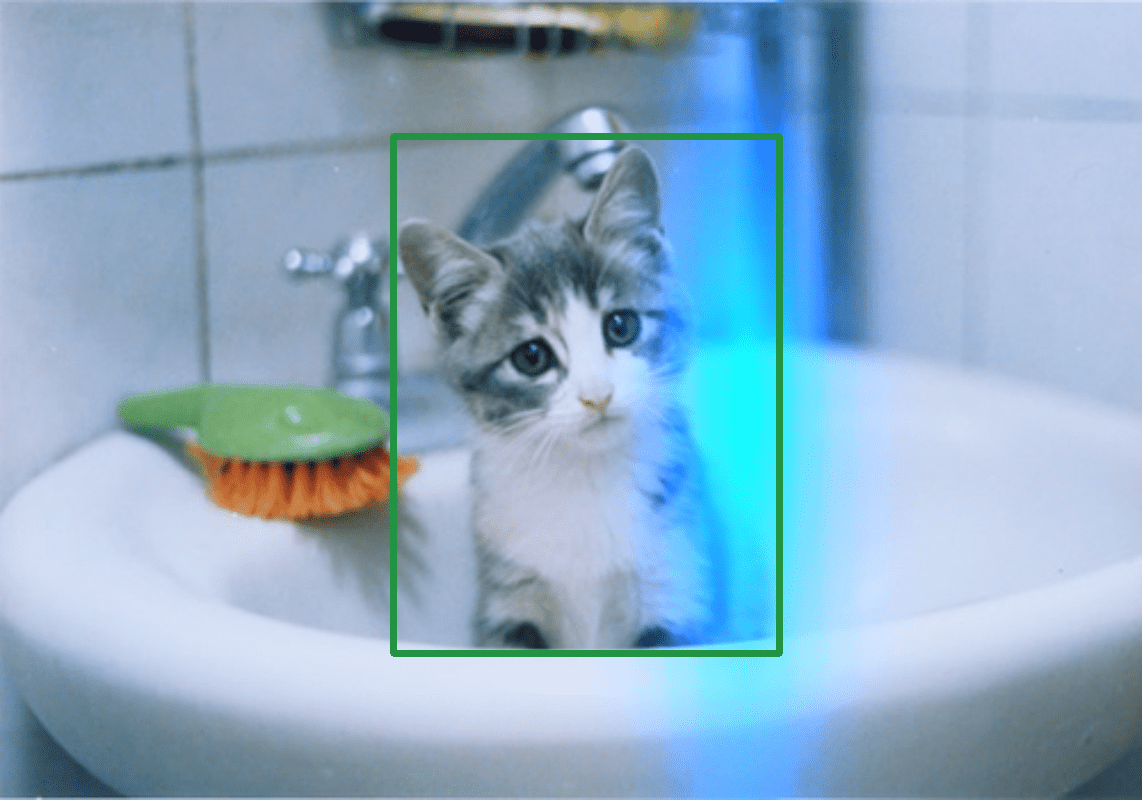}
\includegraphics[ width=0.19\linewidth]{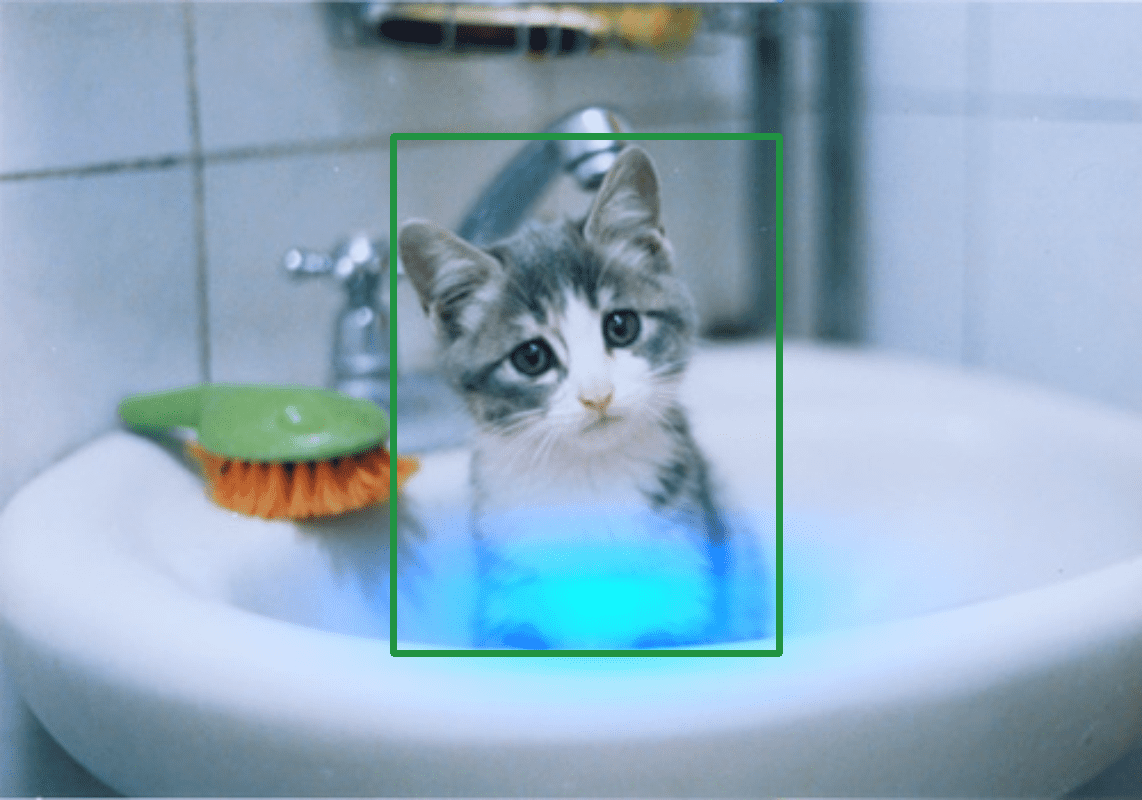}
\includegraphics[ width=0.19\linewidth]{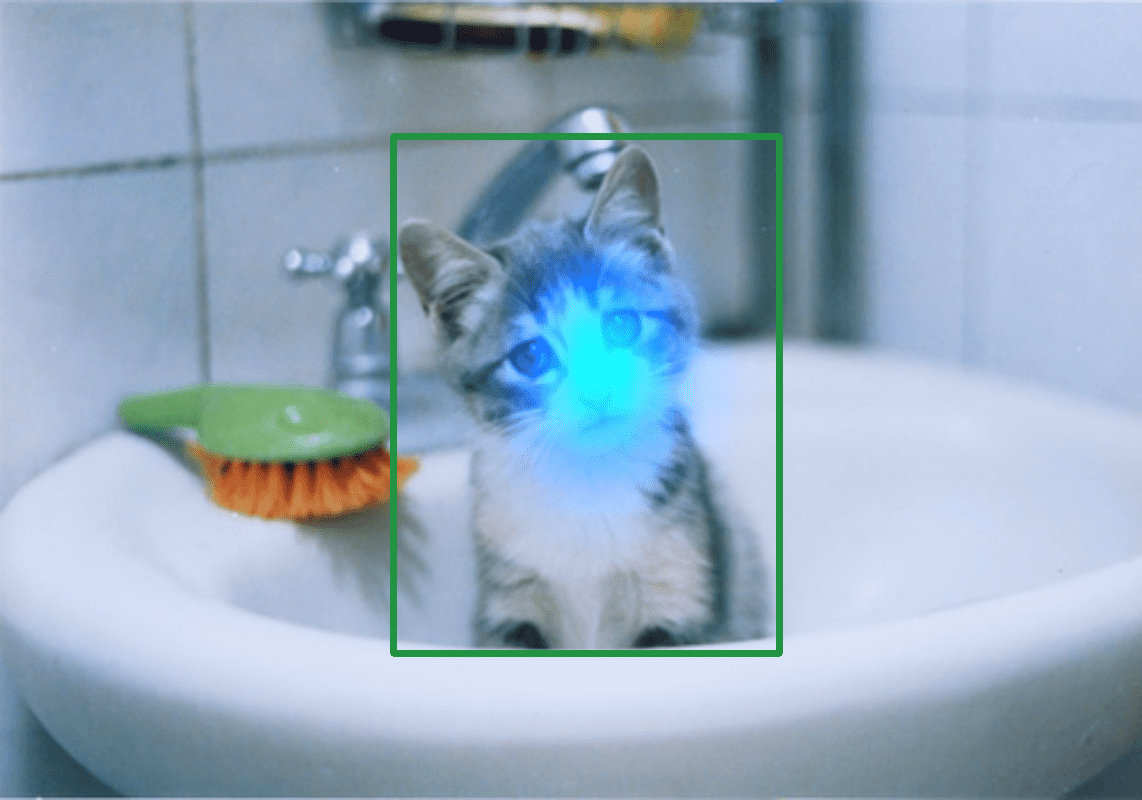}
\\
\includegraphics[ width=0.19\linewidth]{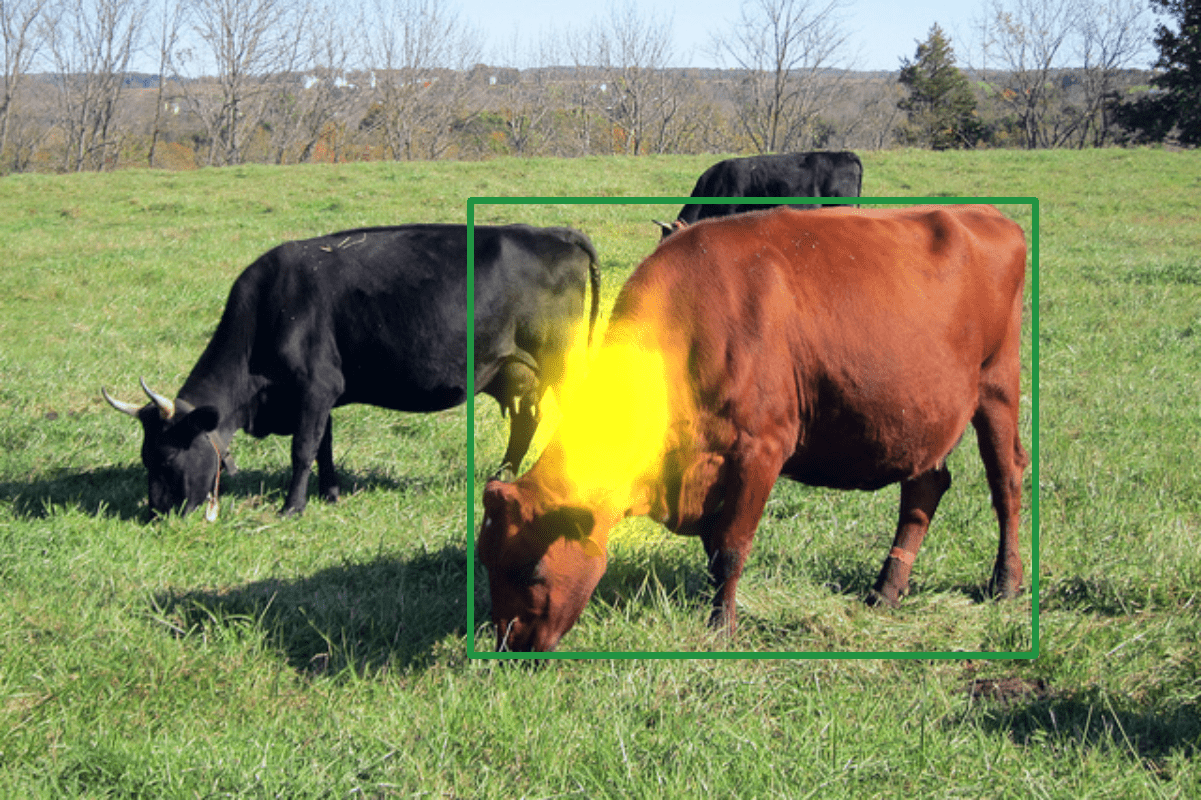}
\includegraphics[ width=0.19\linewidth]{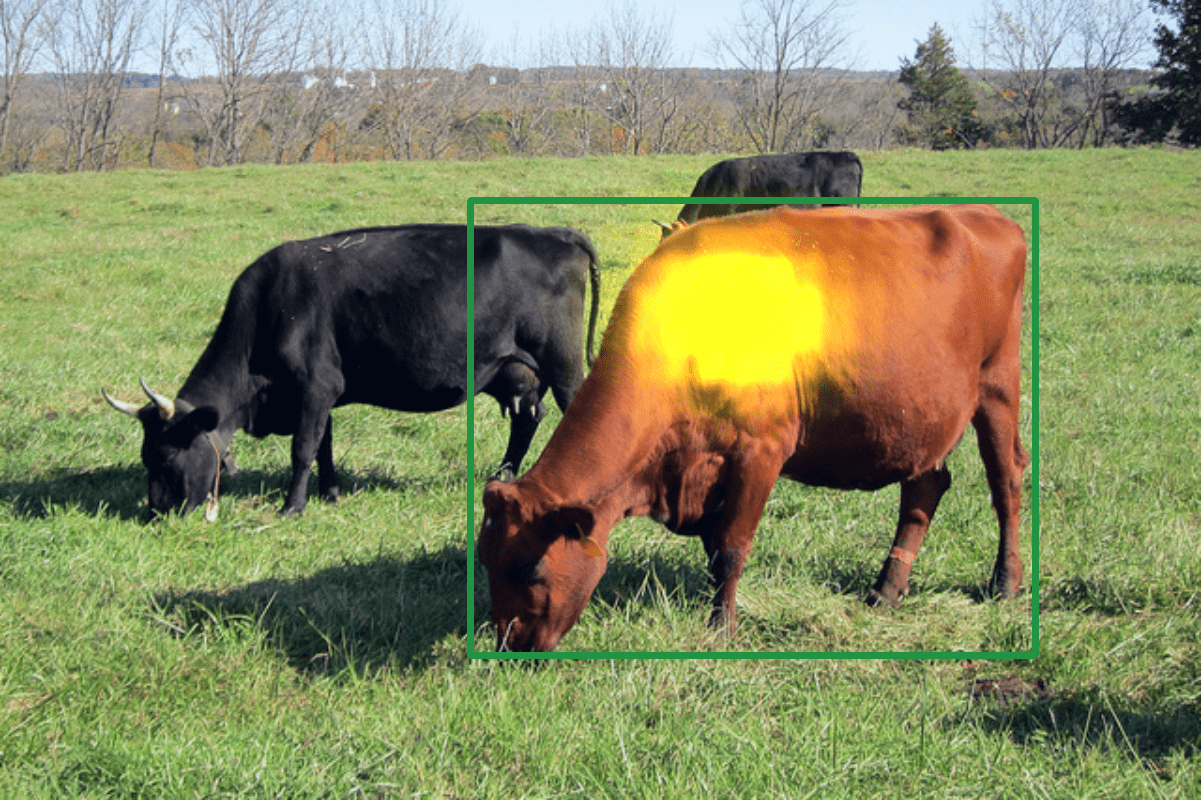}
\includegraphics[ width=0.19\linewidth]{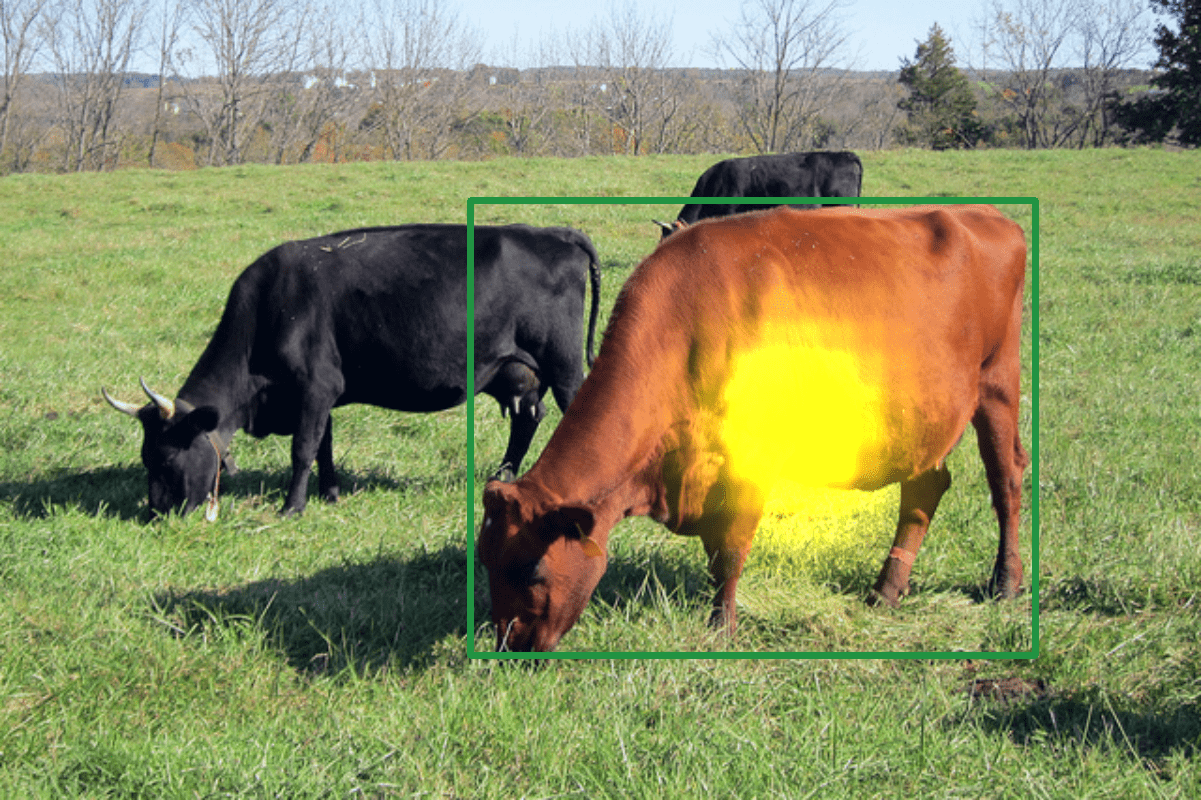}
\includegraphics[ width=0.19\linewidth]{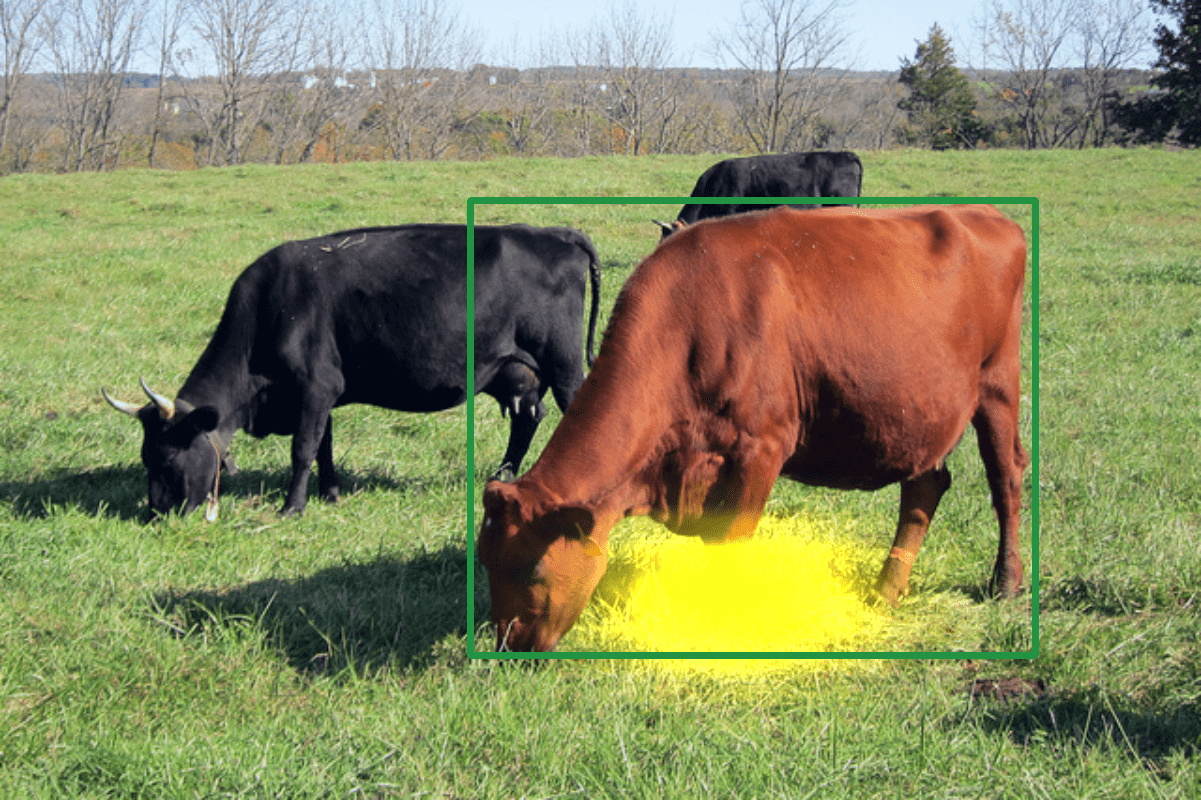}
\includegraphics[ width=0.19\linewidth]{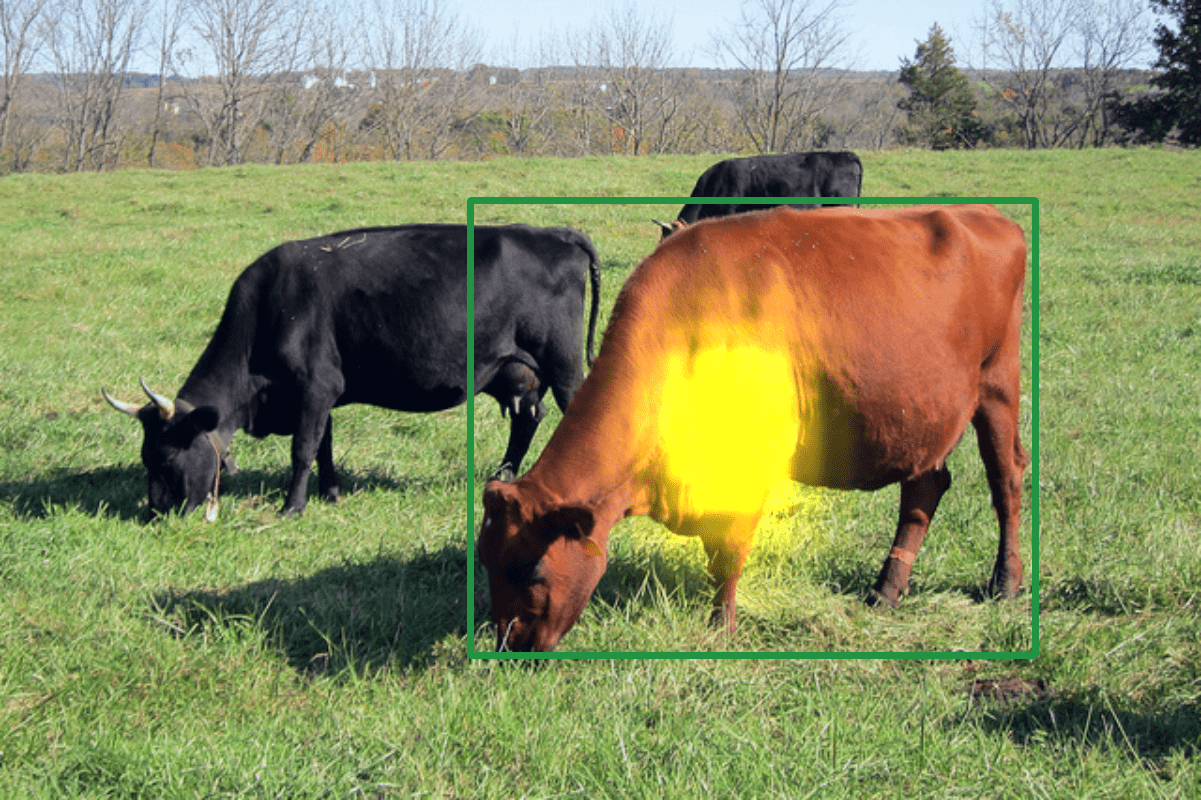}
\\
\includegraphics[ width=0.19\linewidth]{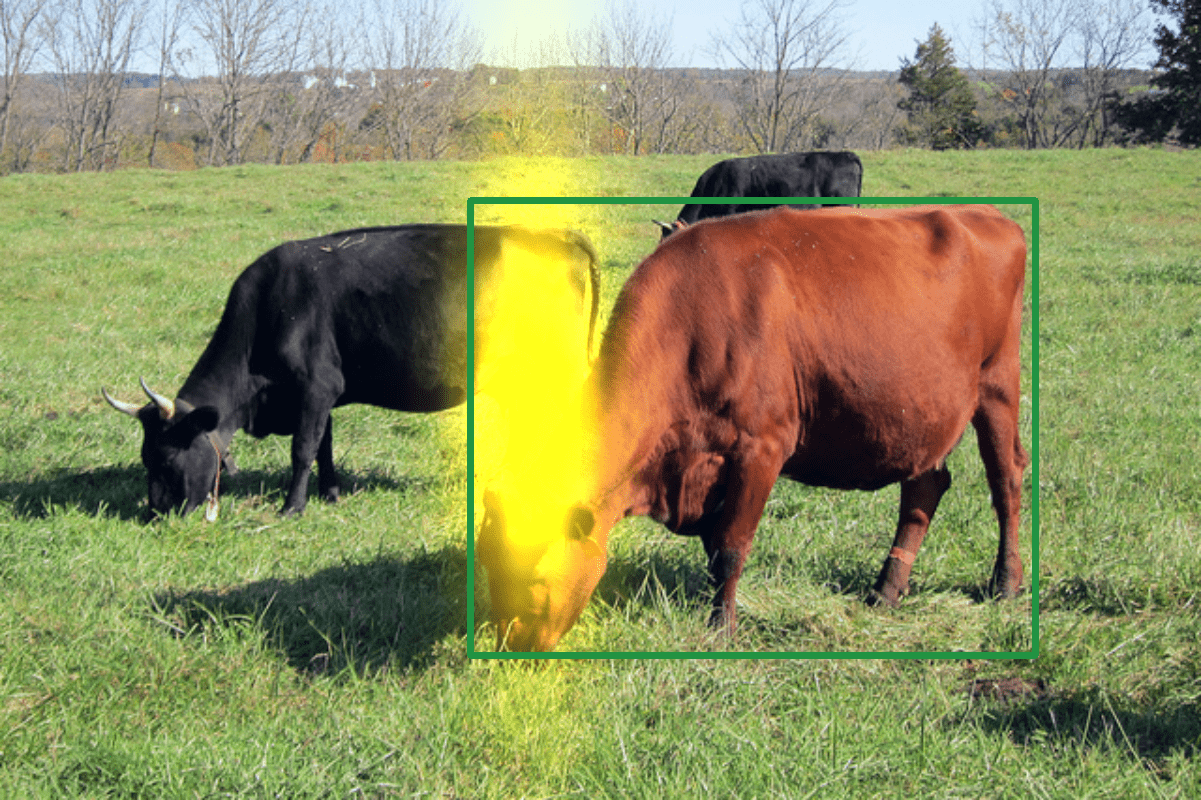}
\includegraphics[ width=0.19\linewidth]{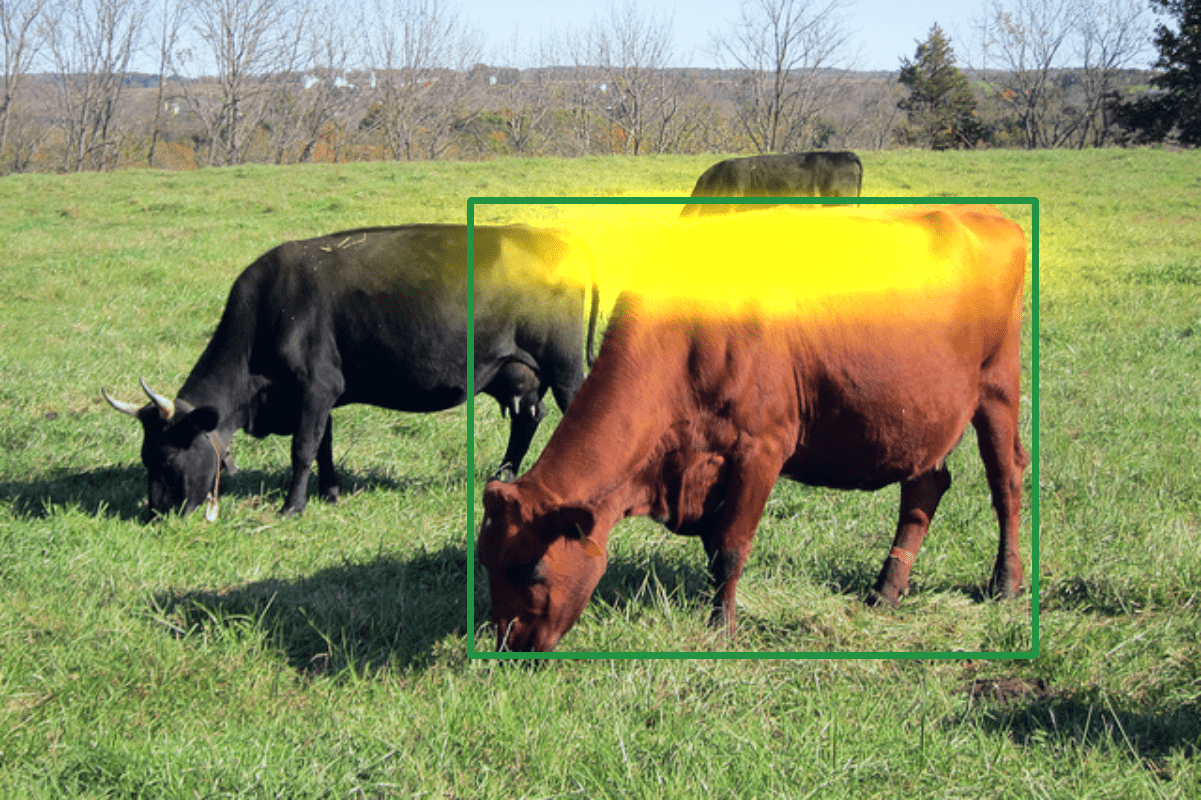}
\includegraphics[ width=0.19\linewidth]{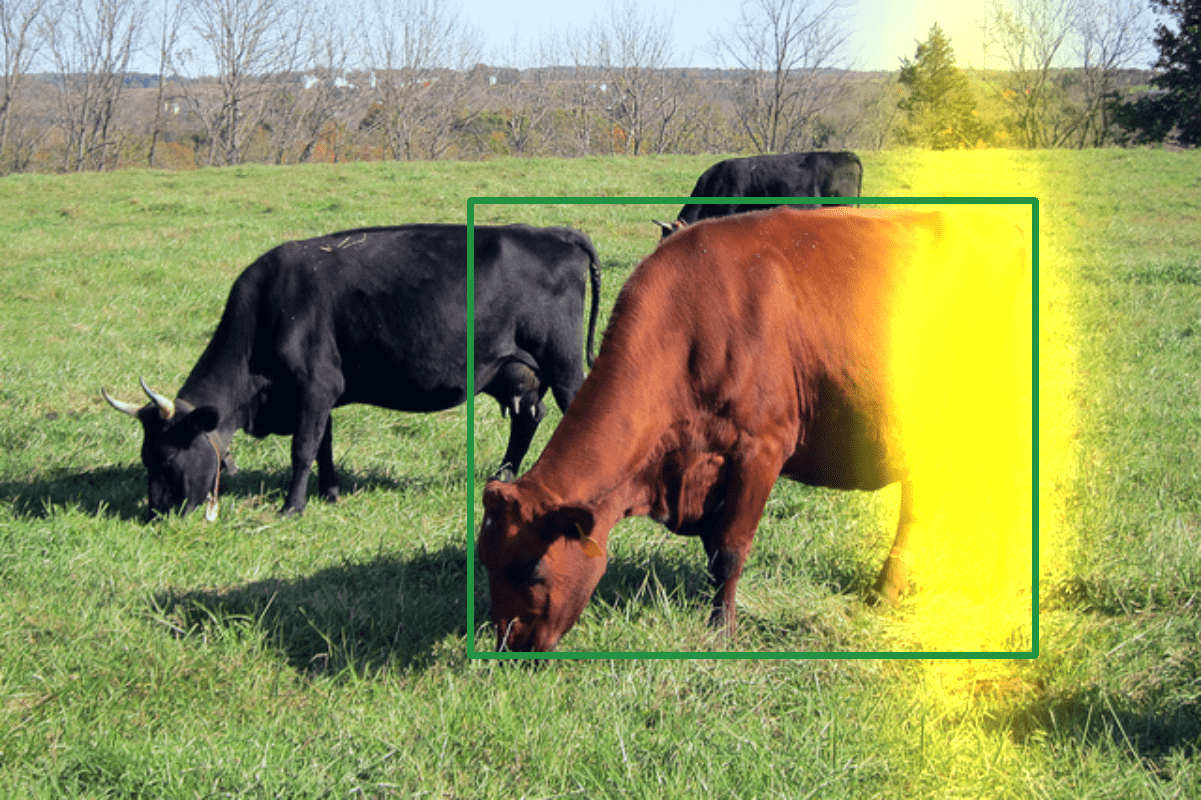}
\includegraphics[ width=0.19\linewidth]{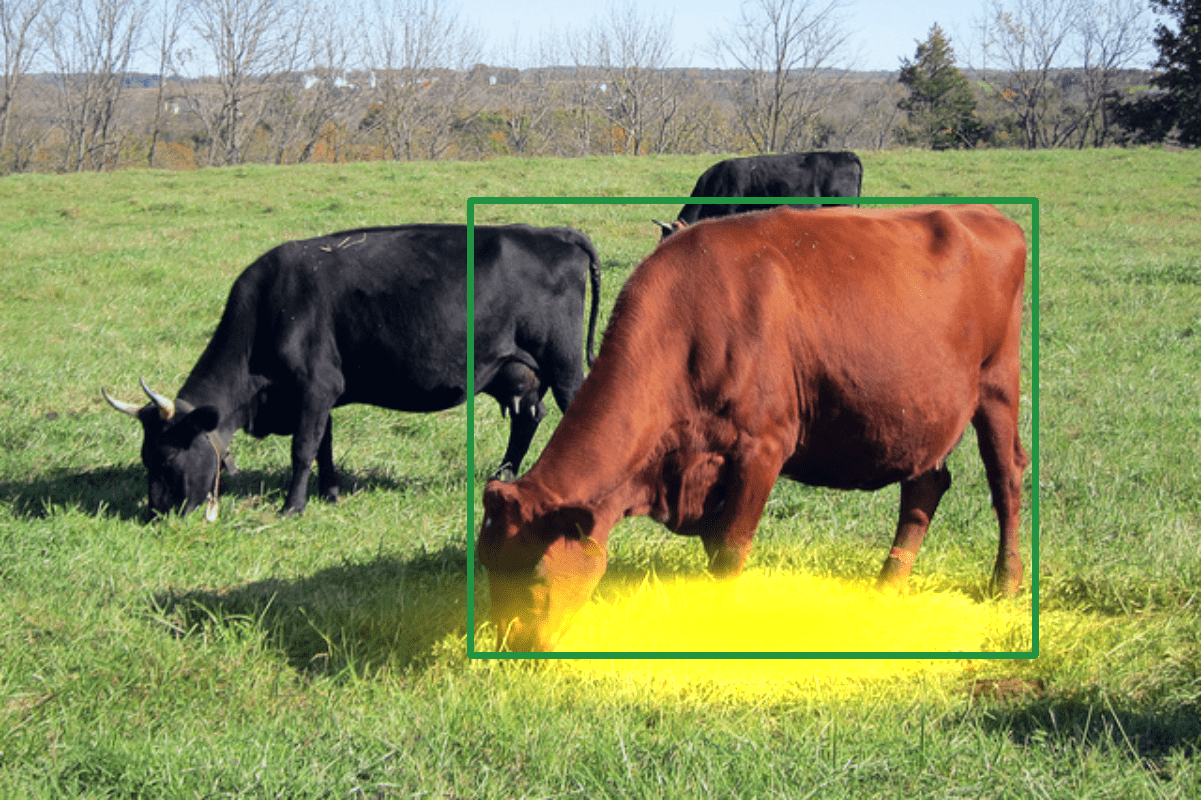}
\includegraphics[ width=0.19\linewidth]{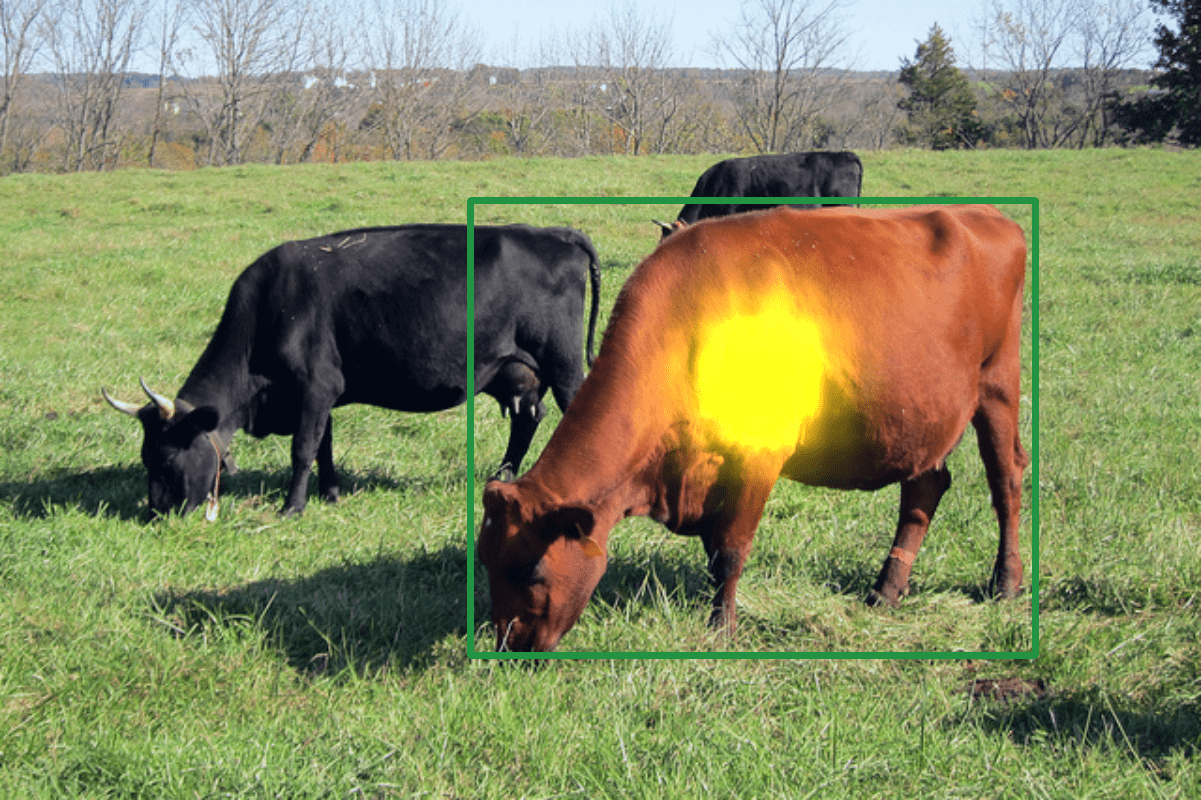}
\\
\includegraphics[ width=0.19\linewidth]{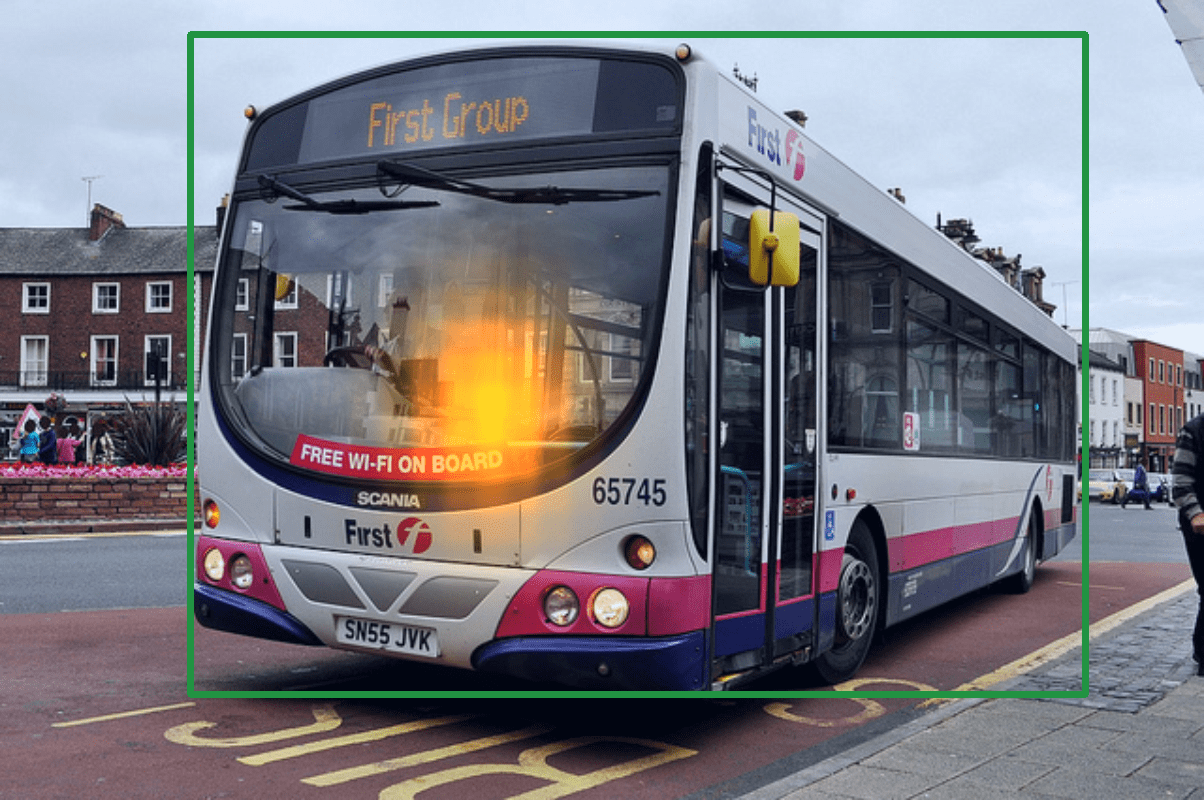}
\includegraphics[ width=0.19\linewidth]{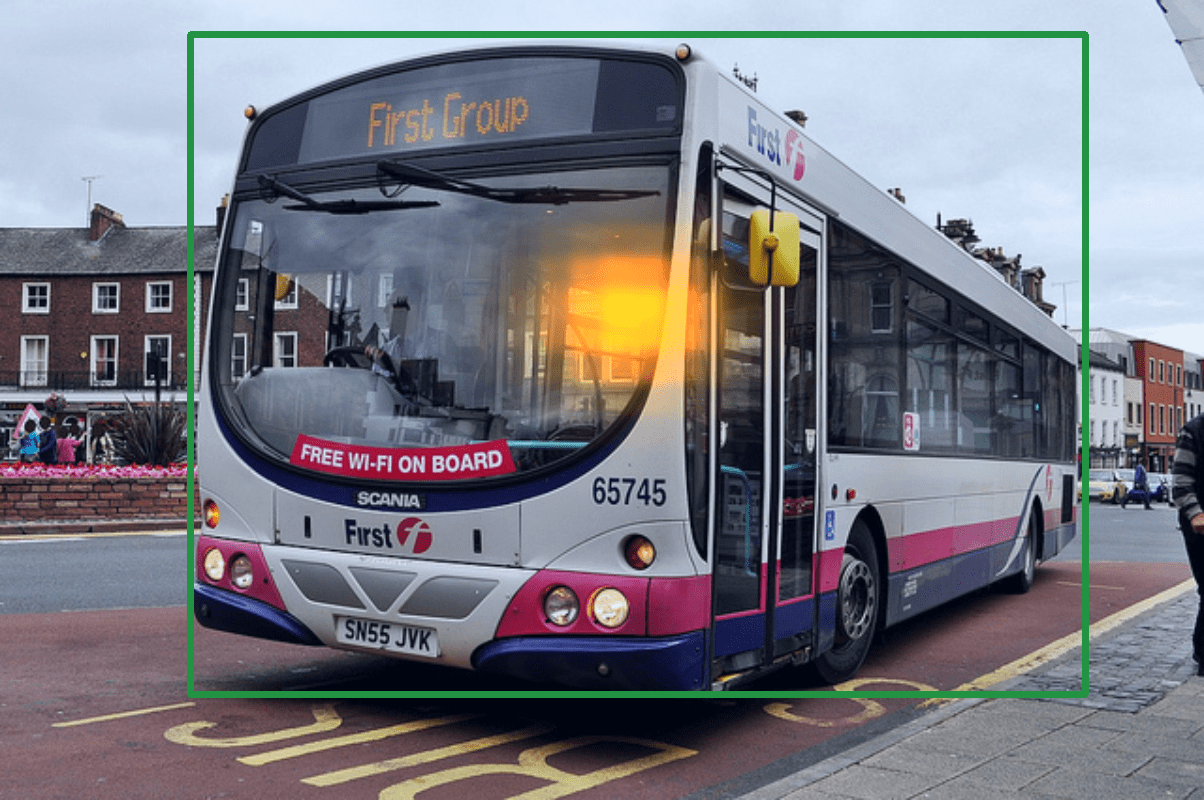}
\includegraphics[ width=0.19\linewidth]{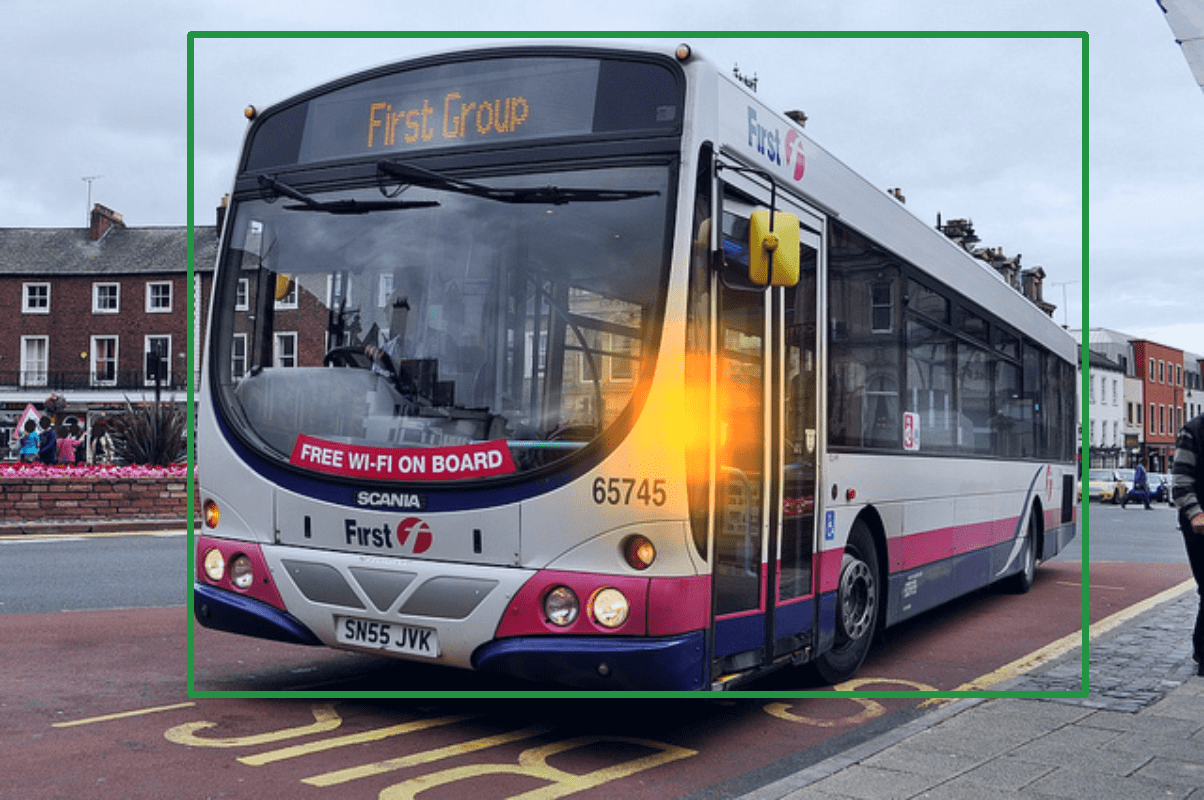}
\includegraphics[ width=0.19\linewidth]{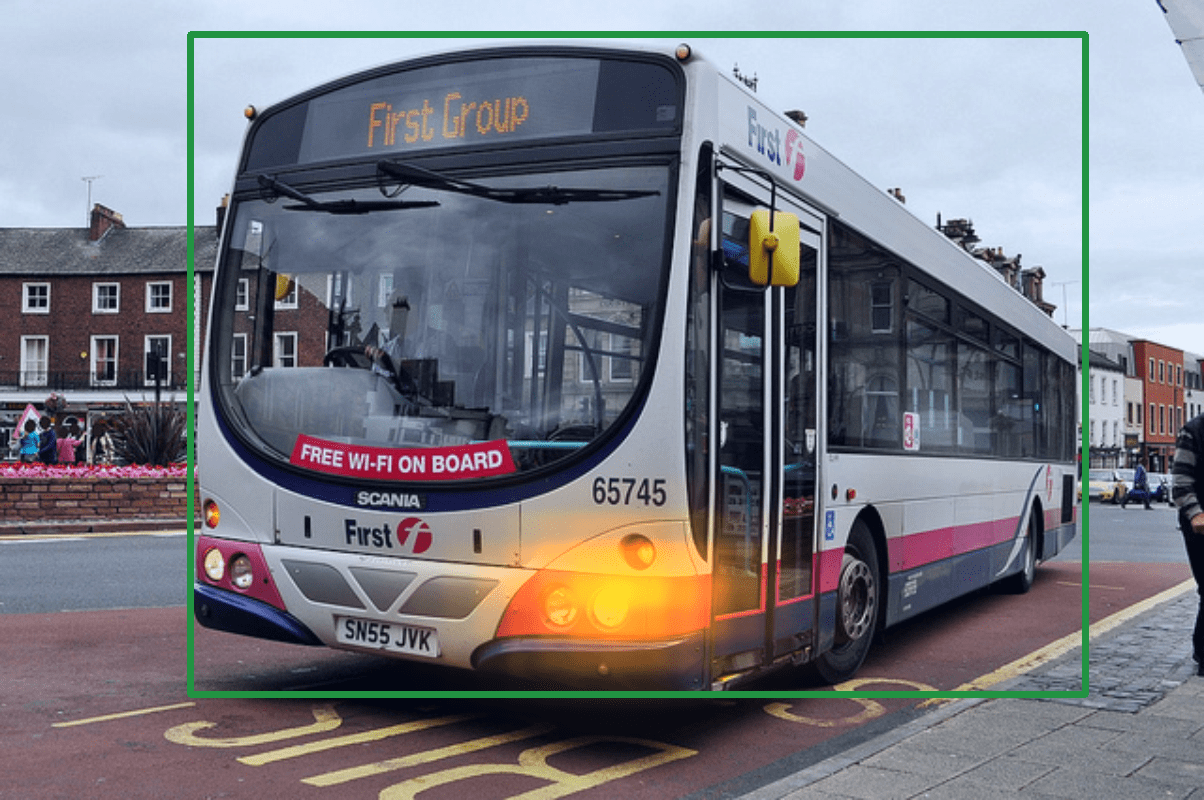}
\includegraphics[ width=0.19\linewidth]{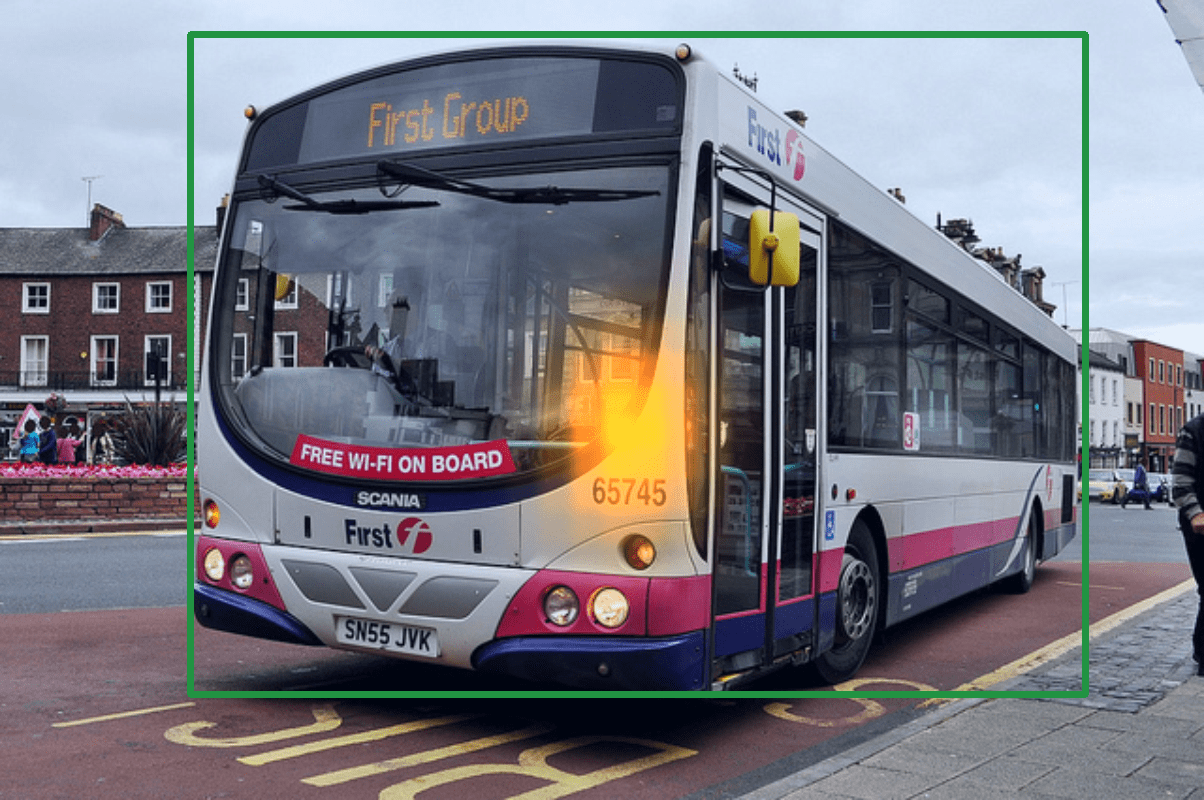}
\\
\includegraphics[ width=0.19\linewidth]{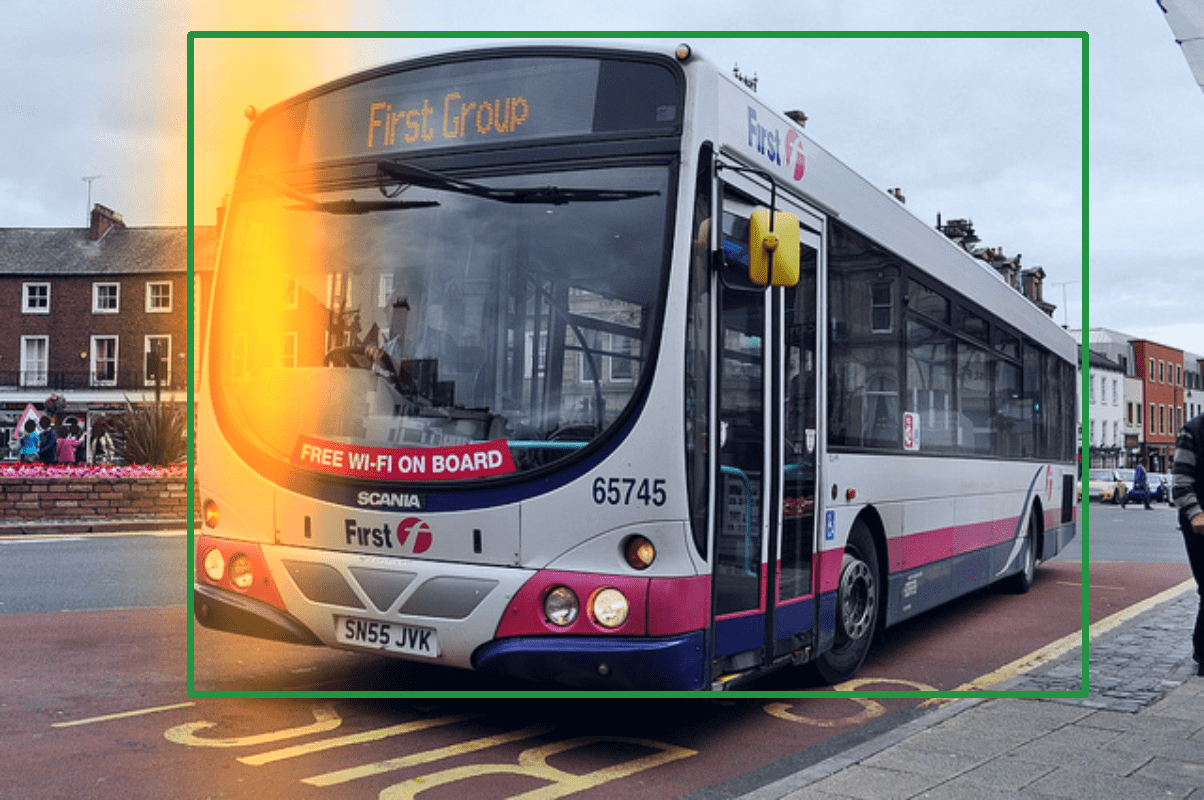}
\includegraphics[ width=0.19\linewidth]{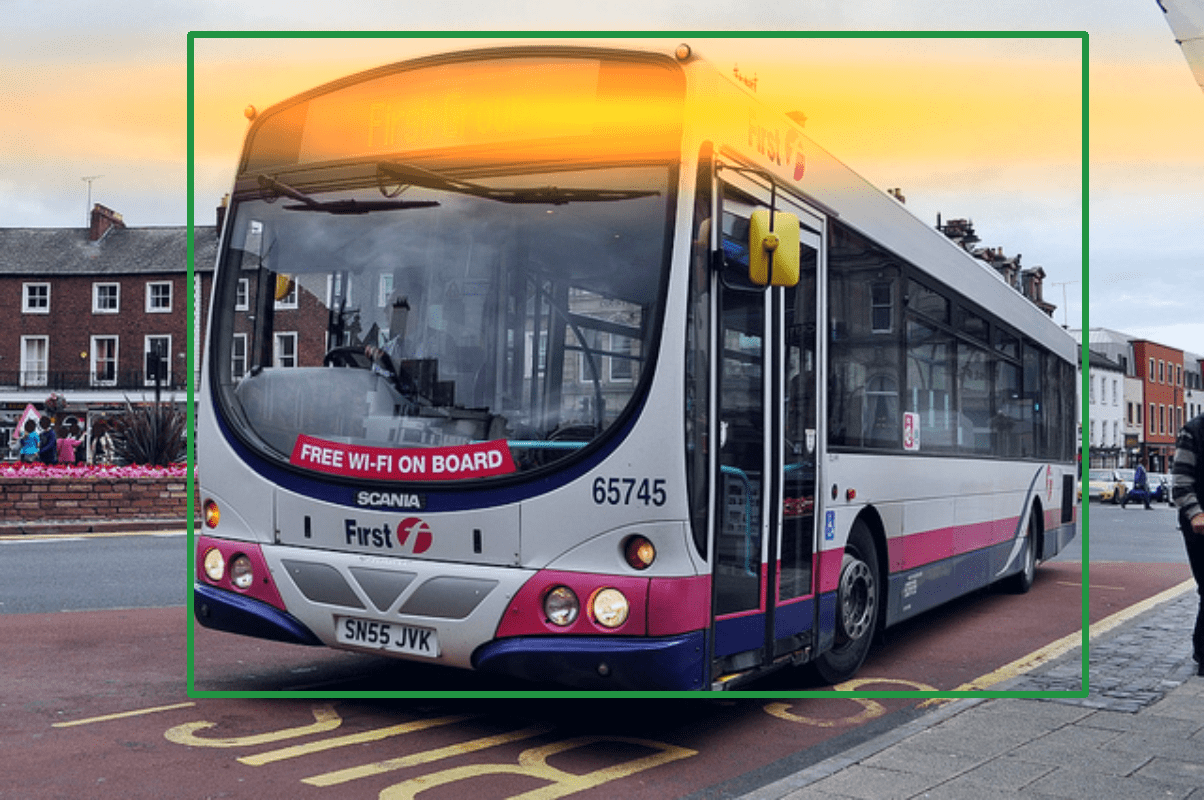}
\includegraphics[ width=0.19\linewidth]{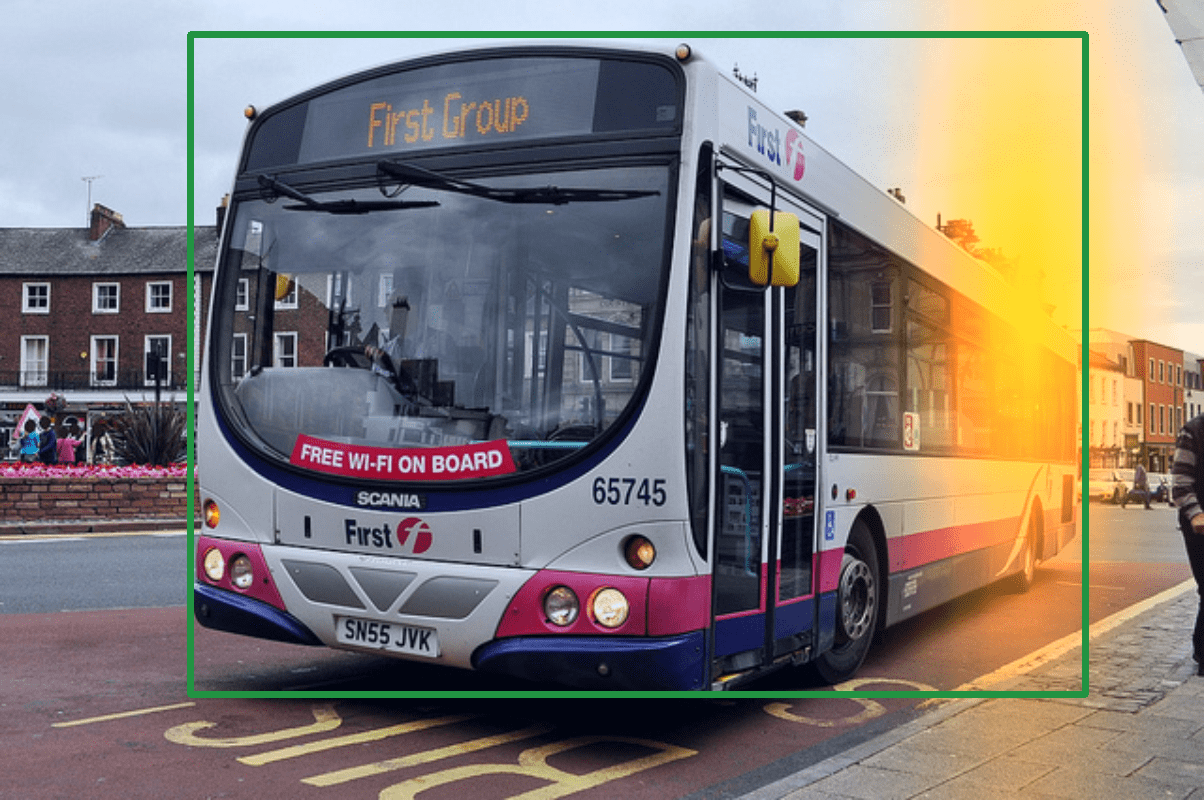}
\includegraphics[ width=0.19\linewidth]{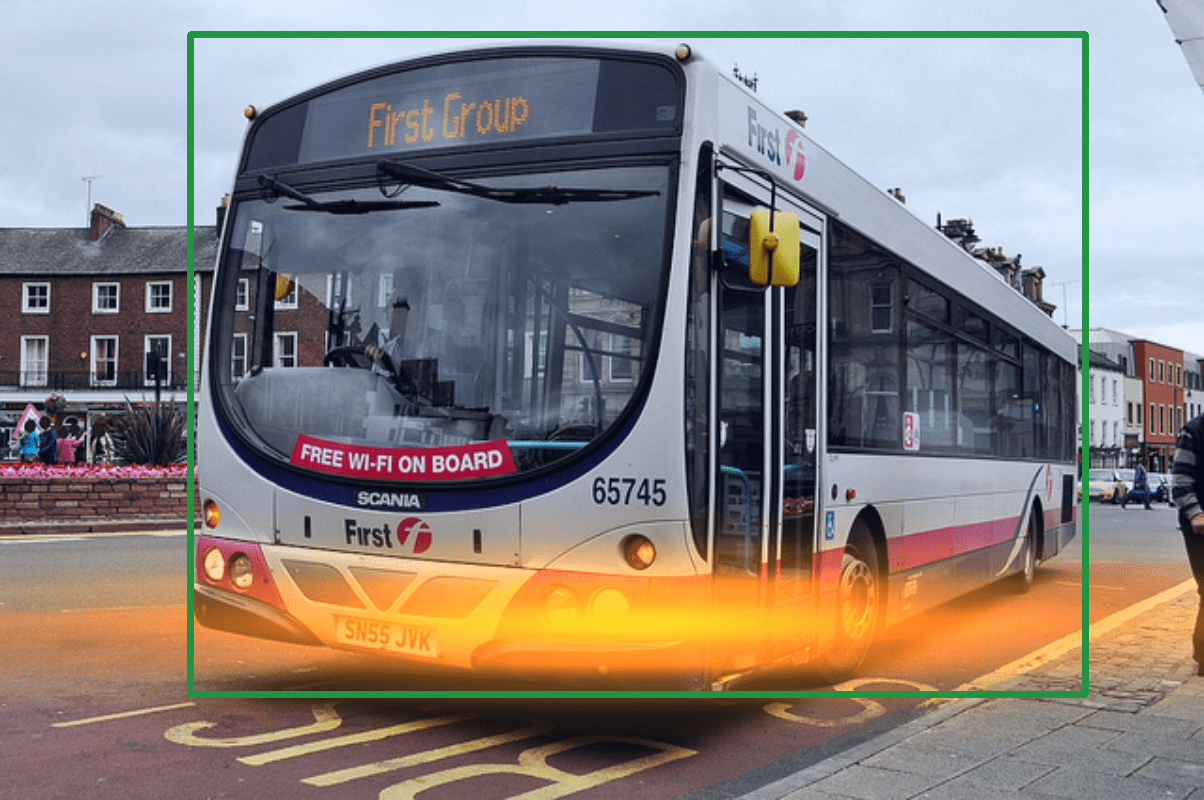}
\includegraphics[ width=0.19\linewidth]{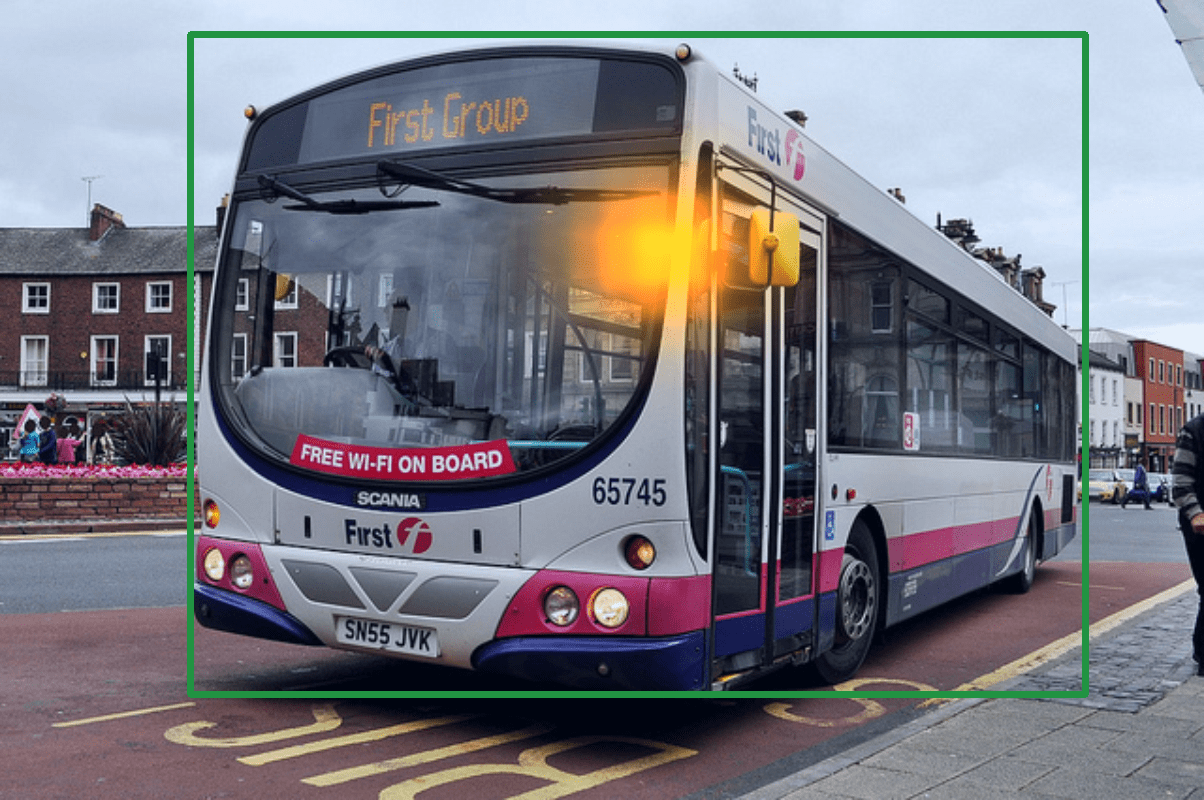}
\\
   \caption{
   Illustrating the spatial attention weight maps from \textbf{the first decoder layer}. For each scene, the upper row is from the model with random initialized box query and the lower row is from the model with image-dependent box query.
   The attention weight maps 
   are from $5$ heads out of the $8$ heads and the green box in each image is the ground-truth box. Best viewed in color.
   }
\label{fig:multiheadattention}
\vspace{-0.5cm}
\end{figure*}

\subsection{Visualization}
Figure~\ref{fig:multiheadattention}
visualizes the spatial attention weight maps of the first decoder layer. The upper row of each scene adopts the randomly initialized box query, and the lower row is our image-dependent box query.
The maps are soft-max normalized 
over the dot products between the box query and the position embedding of the key.
We show $5$ out of the $8$ maps, and the other three are duplicates. The duplicates might be different for models trained several times, but the detection performance is almost the same.

In the visualizations of our method, the spatial attention map in the first four columns highlights an extremity region near a certain edge of the bounding box, $e.g.$, the head of the cat in the second row. The map in the last column highlights a small region inside the box, whose representations might already encode enough information for object classification. We also find that the image-dependent box query is able to scale the spatial spread for the extremity it highlights.
 For a larger object, such as the bus in the last row, our method highlights a larger spread. For a smaller object, such as the cat in the second row, our method highlights a smaller spread.

By contrast, the spatial attention map of the randomly initialized box query does not show this property. Each head only highlights a small region near the object center. Although 
a certain head may try to find an extremity region, such as the fourth column, it could not scale the spatial spread of the highlight regions according to the object size.

\subsection{Discussion: Box Query vs Anchor Box}
The box query builds the connection between
DETR and Faster R-CNN. They both use the form of boxes as one input and extract features according to the box. However, there are some differences.
(1) The box query is in the embedding space, which consists of the embedding of the reference point and the transformation. The transformation could transform the embedding of the reference point to the extremity regions of the object, which is like a box. The anchor box is in the form of $2$D coordinates, consisting of box center and box height/width.
(2)  The box query is explored through the attention mechanism to find the extremities of the object
for box regression and the regions inside the object 
for classification.
While the anchor box is used as the initial guess and all areas in the box are treated as the region of interest (ROI), which is used for classification and box regression.

\subsection{Loss Function}
The box query and the content query are initialized from the top-scored candidate boxes. We supervise the box predictions and view it as a bipartite matching problem between the predicted boxes and the ground-truth boxes. 

The matching cost is defined as:
\begin{align}
    \mathcal{C}_{\operatorname{match}}
    = \sum\nolimits_{i=1}^{G}
    [\mu_{\operatorname{cls}}
    \ell_{\operatorname{FL}}(\mathbf{p}_{\xi(i)}, \bar{c}_i) + \ell_{\operatorname{box}}(\mathbf{b}_{	\xi(i)}, \bar{\mathbf{b}}_i)
    ].
\end{align}
Here $G$ is the number of ground-truth objects. $\ell_{\operatorname{FL}}$ is the focal loss and $\mu_{\operatorname{cls}}=2$ is the trade-off coefficient.
$\bar{c}_i$ is the class of the $i$th ground-truth box and $\bar{c}_i = \operatorname{null}$ if $i > G$. We binarize the ground-truth label and $\bar{c}_i = 1$ if it is an object. 
$\bar{\mathbf{b}}_i$ is the $i$th ground-truth box and $\ell_{\operatorname{box}}$ is a combination of $\ell_1$ loss and GIoU loss~\cite{RezatofighiTGS019} and the loss weights are $5$ and $2$, respectively.

$\xi(\cdot)$ is a permutation of 
N predictions and we could find an optimal solution $\hat \xi(\cdot)$ through minimizing the matching cost.
Then the loss function to supervise the initialization of the box query and the content query is computed as:
\begin{align}
&\mathcal{L}_{init} = \sum_{i=1}^N
\lambda_{\operatorname{cls}}\ell_{\operatorname{FL}}(\mathbf{p}_{\hat \xi(i)}, \bar{c}_i) 
+ 
\delta_{[i \leqslant G]}
    \ell_{
    \operatorname{box}}(\mathbf{b}_{\hat \xi(i)}, \bar{\mathbf{b}}_i),
\end{align}
where $\lambda_{\operatorname{cls}}=2$ is the trade-off coefficient. $\delta$ is the indicator function. We use the same loss function as Conditional DETR~\cite{meng2021conditional} to supervise the output of the decoder, which is noted as $\mathcal{L}_{dec}$. Then the overall loss function is defined as:
\begin{align}
&\mathcal{L} = \mathcal{L}_{init} + \mathcal{L}_{dec}.
\end{align}

\section{Experiments}

\subsection{Setting}
\noindent\textbf{Dataset.}
We perform the experiments on the COCO $2017$~\cite{LinMBHPRDZ14}
detection dataset, which contains about $118$K training (\texttt{train}) images and
$5$K validation (\texttt{val}) images.

\newcommand{\blue}[1]{\textcolor{blue}{#1}}
\begin{table}[h]
        \centering\scriptsize
            \renewcommand{\arraystretch}{1.18}
    \caption{Comparison of Conditional DETR V2 with other detection models on COCO 2017 \texttt{val}.
    $^*$ means these methods use multi-scale feature.
    $^+$ means we use $900$ box queries.
    }
    \vspace{-0.2cm}
    \label{tab:comparisontodetr}
    \resizebox{0.996\columnwidth}{!}{
    \begin{tabular}{lc|cccccccc}
        \shline
        Model & \#epochs & GFLOPs & \#params & AP & AP$_{50}$ & AP$_{75}$ & AP$_{S}$ & AP$_{M}$ & AP$_{L}$\\
        \shline
        {DETR}-R$50$~\cite{CarionMSUKZ20} & $500$ & $86$ & $41$M & $42.0$  & $62.4$  & $44.2$  & $20.5$  & $45.8$  & $61.1$ \\
        {DETR}-R$50$~\cite{CarionMSUKZ20} & $50$ & $86$ & $41$M & $34.9$ &  $55.5$ &  $36.0$ &  $14.4$ &  $37.2$ &  $54.5$ \\
        PNP-DETR-R$50$~\cite{wang2021pnp} & $ 500 $ & $ 82 $ & $ 41 $M & $ 41.8 $ & $ 62.1 $ & $ 44.4 $ & $ 21.2 $ & $ 45.3 $ & $ 60.8 $ \\
         Deformable DETR-R50~\cite{ZhuSLLWD20}  & $50$ & $ 78 $ & $ 34 $M & $ 39.4 $ & $ 59.6 $ & $ 42.3 $ & $ 20.6 $ & $ 42.9 $ & $ 55.5 $ \\
        UP-DETR-R$50$~\cite{DaiCLC20} & $ 150 $ & $ 86 $ & $ 41 $M & $ 40.5 $ & $ 60.8 $ & $ 42.6 $ & $ 19.0 $ & $ 44.4 $ & $ 60.0 $ \\
        Anchor DETR-R50~\cite{wang2021anchor} & $50$ & $ 94 $ & $ 37 $M & $42.1$ & $63.1$ & $44.9$ & $22.3$ & $46.2$ & $60.0$ \\
        Conditional DETR-R50~\cite{meng2021conditional} & $50$ & $90$ & $44$M & $40.9$ & $61.8$ & $43.3$ & $20.8$ & $44.6$ & $59.2$ \\
        \rowcolor{lightgray} Conditional DETR V2-R50 & $50$ & $ 89 $ & $ 46 $M & $ \textbf{42.5} $ & $ 63.4 $ & $ 44.9 $ & $ 22.5 $ & $ 45.9 $ & $ 61.4 $ \\
        \hline
        {DETR}-DC5-R$50$~\cite{CarionMSUKZ20} & $500$ & $187$ & $41$M & $43.3$  & $63.1$  & $45.9$  & $22.5$  & $47.3$  & $61.1$ \\
        {DETR}-DC5-R$50$~\cite{CarionMSUKZ20} & $50$ & $187$ & $41$M & $36.7$ & $57.6$ & $38.2$ & $15.4$ & $39.8$ & $56.3$ \\
        Faster RCNN-FPN-R$50$$^*$~\cite{RenHG017} & $108$ & $180$ & $42$M & $42.0$  & $62.1$ & $45.5$ & $26.6$ & $45.5$ & $53.4$ \\
        TSP-RCNN-R$50$$^*$~\cite{SunCYK20} & $36$ & $188$ & $-$ & $43.8$ & $63.3$ & $48.3$ & $28.6$ & $46.9$ & $55.7$ \\
        SMCA-R$50$$^*$~\cite{GaoZWDL21} & $50$ & $152$ & $40$M & $43.7$ & $63.6$ & $47.2$ & $24.2$ & $47.0$ & $60.4$ \\
        PNP-DETR-DC5-R$50$~\cite{wang2021pnp} & $ 500 $ & $ 144 $ & $ 41 $M & $ 43.1 $ & $ 63.4 $ & $ 45.3 $ & $ 22.7 $ & $ 46.5 $ & $ 61.1 $ \\
        Deformable DETR-DC5-R50~\cite{ZhuSLLWD20}  & $ $50$ $ & $ 128 $ & $ 34 $M & $ 41.5 $ & $ 61.8 $ & $ 44.9 $ & $ 24.1 $ & $ 45.3 $ & $ 56.0 $ \\
        Two-Stage Deformable DETR-DC5-R50~\cite{ZhuSLLWD20}  & $ $50$ $ & $ 128 $ & $ 34 $M & $ 43.6 $ & $ 63.6 $ & $ 47.0 $ & $ 25.8 $ & $ 46.6 $ & $ 58.1 $ \\
        Anchor DETR-DC5-R50~\cite{wang2021anchor} & $50$ & $ 173 $ & $ 37 $M & $44.2$ & $64.7$ & $47.5$ & $24.7$ & $48.2$ & $60.6$ \\
        Conditional DETR-DC5-R50\cite{meng2021conditional} & $50$ & $195$ & $44$M &  $43.8$ & $64.4$ &  $46.7$ &  $24.0$ & $47.6$ &  $60.7$  \\
        \rowcolor{lightgray} Conditional DETR V2-DC5-R50 & $50$ & $ 161 $ & $ 46 $M & $ 44.8 $ & $ 65.3 $ & $ 48.2 $ & $ 25.5 $ & $ 48.6 $ & $ 62.0 $ \\
        \rowcolor{lightgray} Conditional DETR V2-DC5-R50$^+$ & $50$ & $ 181 $ & $ 46 $M & $ \textbf{45.2} $ & $ 66.0 $ & $ 48.4 $ & $ 26.5 $ & $ 49.0 $ & $ 62.1 $ \\
        \hline
        {DETR}-R$101$~\cite{CarionMSUKZ20} & $500$ & $152$ & $60$M & $43.5$  & $63.8$  & $46.4$  & $21.9$  & $48.0$  & $61.8$ \\
        {DETR}-R$101$~\cite{CarionMSUKZ20} & $50$ & $152$ & $60$M & $36.9$ & $57.8$ & $38.6$ & $15.5$ & $40.6$ & $55.6$ \\
        Anchor DETR-R101~\cite{wang2021anchor} & $50$ & $ 160 $ & $ 56 $M & $43.5$ & $64.3$ & $46.6$ & $23.2$ & $47.7$ & $61.4$ \\
        Conditional DETR-R101~\cite{meng2021conditional}  & $50$ & $156$ & $63$M  & $42.8$  & $63.7$  & $46.0$  & $21.7$ & $46.6$ & $60.9$ \\
        \rowcolor{lightgray} Conditional DETR V2-R101 & $50$ & $ 155 $ & $ 65 $M & $ \textbf{43.9} $ & $ 65.3 $ & $ 46.7 $ & $ 25.2 $ & $ 48.0 $ & $ 62.6 $ \\
        \hline
        DETR-DC5-R$101$~\cite{CarionMSUKZ20}  & $ 500 $ & $ 253 $ & $ 60 $M & $ {44.9}  $ & $ {64.7}  $ & $ 47.7  $ & $ 23.7  $ & $ {49.5}  $ & $ {62.3} $\\
        DETR-DC5-R$101$~\cite{CarionMSUKZ20}  & $ $50$ $ & $ 253 $ & $ 60 $M & $ 38.6 $ & $ 59.7 $ & $ 40.7 $ & $ 17.2 $ & $ 42.2 $ & $ 57.4 $\\
        Faster RCNN-FPN-R$101$$^*$~\cite{RenHG017} & $108$ & $246$ & $60$M & $44.0$ & $63.9$ & $47.8$ & $27.2$ & $48.1$ & $56.0$ \\
        TSP-RCNN-R$101$$^*$~\cite{SunCYK20} & $36$ & $254$ & $-$ & $44.8$ & $63.8$ & $49.2$ & $29.0$ & $47.9$ & $57.1$ \\
        SMCA-R$101$$^*$~\cite{GaoZWDL21} & $50$ & $218$ & $58$M & $44.4$ & $65.2$ & $48.0$ & $24.3$ & $48.5$ & $61.0$  \\
        Anchor DETR-DC5-R101~\cite{wang2021anchor} & $50$ & $ 239 $ & $ 56 $M & $45.1$ & $65.7$ & $48.8$ & $25.8$ & $49.4$ & $61.6$ \\
        Conditional DETR-DC5-R101\cite{meng2021conditional} & $50$ & $262$ & $63$M & $45.0$ &  $65.5$ & $48.4$ & $26.1$ & $48.9$ & $62.8$  \\
        \rowcolor{lightgray} Conditional DETR V2-DC5-R101 & $50$ & $ 228 $ & $ 65 $M & $ 45.5 $ & $ 66.0 $ & $ 48.7 $ & $ 26.3 $ & $ 49.0 $ & $ 62.9 $ \\
        \rowcolor{lightgray} Conditional DETR V2-DC5-R101$^+$ & $50$ & $ 247 $ & $ 65 $M & $ \textbf{45.9} $ & $ 66.6 $ & $ 49.6 $ & $ 27.4 $ & $ 49.9 $ & $ 62.6 $ \\
        \hline
        Conditional DETR-HR48~\cite{meng2021conditional} & $50$ & $ 1090 $ & $ 87 $M & $ 48.2 $ & $ 68.2 $ & $ 52.4 $ & $ 30.6 $ & $ 52.3 $ & $ 64.3 $ \\
        \rowcolor{lightgray} Conditional DETR V2-HR48 & $50$ & $ 521 $ & $ 90 $M & $ \textbf{49.8} $ & $ 70.2 $ & $ 54.2 $ & $ 32.1 $ & $ 53.3 $ & $ 65.9 $ \\
        \shline
    \end{tabular}
    }
\end{table}

\vspace{.1cm}
\noindent\textbf{Training.}
We follow the DETR training protocol~\cite{CarionMSUKZ20}.
The backbone is the ImageNet-pretrained model
from TORCHVISION with batchnorm layers fixed,
and the transformer parameters are initialized
using the Xavier initialization scheme~\cite{GlorotB10}.
We train the model on the COCO training set for $50$ epochs, with the AdamW~\cite{LoshchilovH17} optimizer.
The weight decay is set to be $10^{-4}$.
The learning rates for the backbone and the transformer
are initialized as
$10^{-5}$ and $10^{-4}$, respectively. The learning rate is dropped by a factor of $10$
after $40$ epochs.
The dropout rate in the transformer is $0.1$.
The head for the attention is $8$, the attention feature channel is $256$ and the hidden dimension of the feed-forward network is $2048$. The number of box queries is set as $300$ by default.

We use the augmentation scheme same as DETR~\cite{CarionMSUKZ20}:
resize the input image such that
the short side is at least $480$
and at most $800$ pixels
and the long side is at most $1333$ pixels; 
randomly crop the image such that
a training image is cropped with a probability $0.5$
to a random rectangular patch.

\vspace{.1cm}
\noindent\textbf{Evaluation.}
We use the standard COCO evaluation.
We report the average precision (AP),
and the AP scores at $0.50$, $0.75$
and for the small, medium, and large objects.

\renewcommand{\baselinestretch}{0.975}
\selectfont

\subsection{Results}

Table~\ref{tab:comparisontodetr} shows comparison of the proposed method with other detection methods on COCO 2017 \texttt{val}.
We first report the results of single scale methods:
DETR~\cite{CarionMSUKZ20}, PNP-DETR~\cite{wang2021pnp},
deformable DETR~\cite{ZhuSLLWD20}, UP-DETR~\cite{DaiCLC20}, Anchor DETR~\cite{wang2021anchor} and Conditional DETR~\cite{meng2021conditional}.
We follow~\cite{CarionMSUKZ20} and report the results over
four backbones:
ResNet-$50$~\cite{HeZRS16},
ResNet-$101$,
and their $16\times$-resolution extensions
DC$5$-ResNet-$50$
and
DC$5$-ResNet-$101$.
The $16\times$-resolution extensions are obtained by adding a dilation convolution in the last stage and removing a stride from the first convolution of the last stage.
Our Conditional DETR V2 with $50$ training epochs for R$50$ and R$101$ as the backbones not only outperforms Conditional DETR, but also outperforms DETR with $500$ training epochs.
When it comes to a higher-resolution backbone such as DC$5$-ResNet-$50$, our Conditional DETR V2 saves $17$\% GFLOPs and improves AP score by $1.0$ compared to Conditional DETR. 

In addition, we list some methods that use the multi-scale feature. Although our main purpose is not to study how to make better use of the multi-scale feature, our single-scale models still achieve similar performance to most of theirs. In the meanwhile, our computational cost ($e.g.,$ GFLOPs) is lower than most of these methods and comparable to SMCA, which illustrates the effectiveness of the proposed method.

We also adopt HRNet-W$48$ as the backbone to verify the generalization ability of the proposed method. We down-sample the feature map from $4 \times$ (the initial output of HRNet) to $8 \times$ to save the computation cost. Results in Table~\ref{tab:comparisontodetr} show that the proposed method outperforms Conditional DETR by $1.6$ AP while saving $52$\% GFLOPs. This illustrates that the proposed method could generalize well to the high-resolution backbone.

\subsection{Ablations}
\noindent\textbf{The effect of box query initialization.} We empirically study how the model benefits from the image-dependent box query. 
We compare three ways to initialize the box query:
(i) RB - we use the box queries
to replace the object queries
that Conditional DETR uses in the first decoder layer. The reference point and the transformation ($\mathbf{\lambda}_q$) are randomly initialized.
(ii) IP - the reference point is initialized from the image content and the $\mathbf{\lambda}_q$ is randomly initialized. This is similar to the two-stage strategy in Deformable DETR. 
(iii) IP+IT - both the reference point and the $\mathbf{\lambda}_q$ are initialized from the image content.

\begin{table}[hpt]
    \centering
    \setlength{\tabcolsep}{10pt}
        \scriptsize
            \renewcommand{\arraystretch}{1.25}
    \caption{\small The empirical results 
    about the box query. 
    RB = random box query. IP = image-dependent reference point. IT = image-dependent transformation. The backbone ResNet-50 is adopted. The Horizontal-Vertical Attention is not used here.}
    \label{tab:abla-box-query}
    \begin{tabular}{ l  c  c}
    \shline
         Method & GFLOPs & AP \\
         \shline
         Conditional DETR & $89.5$ & $40.9$   \\
         Conditional DETR + RB & $89.5$ & $40.9$   \\
         Conditional DETR + IP & $90.0$ & $41.9$  \\
         \rowcolor{lightgray} Conditional DETR + IP + IT & $90.1$ & $42.9$  \\
        \shline
    \end{tabular}
\vspace{-0.3cm}
\end{table}

\begin{table}[ht]
    \centering
    \setlength{\tabcolsep}{8pt}
        \scriptsize
            \renewcommand{\arraystretch}{1.25}
    \caption{\small The empirical results 
    about the choice of $[c_w~c_h]$. The backbone ResNet-50 is adopted. The Horizontal-Vertical Attention is not used here.}
    \label{tab:abla-anchor}
    \begin{tabular}{ c c c c c}
    \shline
         $[0.1,0.1]$ & $[0.2,0.2]$ & $[0.4,0.4]$ & GFLOPs &AP  \\
         \shline
         \cmark &  &  &  $89.8$ & $42.2$ \\
          & \cmark  &  &  $89.8$ & $42.2$ \\
          &  & \cmark &  $89.8$ & $42.3$ \\
          \cmark & \cmark  &   &  $90.0$ & $42.6$ \\
         \rowcolor{lightgray} \cmark & \cmark  & \cmark &  $90.1$ & $42.9$ \\
         
        \shline
    \end{tabular}
    \vspace{-0.2cm}
\end{table}

\begin{table}[ht]
    \centering
    \setlength{\tabcolsep}{9pt}
        \scriptsize
            \renewcommand{\arraystretch}{1.25}
    \caption{\small The empirical results 
    about the effect of $[c_x~c_y]$ and $[c_w~c_h]$ that used for content query initialization. The backbone ResNet-50 is adopted. The Horizontal-Vertical Attention is not used here.}
    \label{tab:abla-xyhw}
    \begin{tabular}{ c c c c c}
    \shline
         $[c_x~c_y]$ & $[c_w~c_h]$ &  GFLOPs &AP \\
         \shline
         \cmark &  &  $ 89.8 $ & $ 42.0 $ \\
          & \cmark &  $ 89.8 $ & $ 42.4 $ \\
         \rowcolor{lightgray} \cmark & \cmark &  $90.1$ & $42.9$ \\
         
        \shline
    \end{tabular}
\end{table}

Results are reported in Table~\ref{tab:abla-box-query}. We find the randomly initialized box query does not improve the performance, since the query has no image-dependent information. If we initialize the reference point with positions selected from the image content, the AP improves $1.0$. This is reasonable because the initial reference points are likely in the object regions that we want to pay attention to. If we further predict the transformation ($\lambda_q$) of the box query from the 
image content as well, which contains the scale information of the object, the performance improves to $42.9$ AP. This indicates that both the initializations of the reference point and the transformation in the box query are important.
With an appropriate transformation that contains scale information, we could dynamically adjust the spread of the attended regions for different objects.

\noindent\textbf{The effect of content query initialization.}
We initialize the content query with the estimated box, as  illustrated in Eq.~(\ref{eq:content-query-init}).
We first conduct experiments to verify how different choices of $[c_w~c_h]$ affect the performance. In implementation, we estimate three candidate boxes with different scales at each position and the $[c_w~c_h]$ are [$0.1, 0.1$], [$0.2, 0.2$] and [$0.4, 0.4$], respectively. Please note that $0.1$ is the normalized length, which is the ratio of box length to image length. In Table~\ref{tab:abla-anchor}, we try different choices, $e.g.,$ estimate one or two boxes at one position. Results show that with only one candidate box per position, the performance has already improved to $42.2$ AP, which is $1.3$ higher than Conditional DETR. When we estimate candidate boxes of two scales, the performance further improves. Three scales performs the best. This is because the model's ability to detect objects of different scales is stronger.

When $[c_w, c_h]$ is set as $[0.1,0.1], [0.2,0.2], [0.4,0.4]$, the AP scores for large objects are $61.3, 61.3, 62.0$, and the AP scores for small objects are $22.3, 22.0, 21.9$. With a larger initial scale, large objects will be easier to detect. The performance of small objects drops a little because it is harder to search for small objects with a too large initial scale.

Then, we verify the effect of $[c_x~c_y]$ and $[c_w~c_h]$ for content query initialization in Table~\ref{tab:abla-xyhw}. $[c_x~c_y]$ helps search the distinct region of the object for classification and $[c_w~c_h]$ helps search the region with similar scale for box regression. Using the two items together achieves the overall best result.

\begin{table}[ht]
    \centering
    \setlength{\tabcolsep}{6.5pt}
        \scriptsize
            \renewcommand{\arraystretch}{1.3}
    \caption{\small The analysis of the computational cost and the speed. HVA = Horizontal-Vertical Attention.  The backbone ResNet-$50$ and DC$5$-ResNet-$50$ are adopted. We follow DETR to test over the first 100 images of COCO 2017 \texttt{val}. The experiments are conducted on a single Tesla V100 GPU.}
    \vspace{-0.1cm}
    \label{tab:abla-compute-cost}
    \begin{tabular}{ l  l c c c c}
    \shline
         Method & Backbone & GFLOPs & Memory (M) & FPS & AP \\
         \shline
         Conditional DETR & ResNet-$50$ & $90$ & $2887$ & $22$ & $40.9$   \\
         Conditional DETR v2 w/o HVA & ResNet-$50$ & $ 90 $ & $2913$ & $20$ & $42.9$   \\
         \rowcolor{lightgray} Conditional DETR v2 & ResNet-$50$ & $ 89 $ & $2093$ & $20$ & $42.5$   \\
         \hline
         Conditional DETR & DC$5$-ResNet-$50$ & $ 195 $ & $18631$ & $10$ & $ 43.8 $   \\
         Conditional DETR v2 w/o HVA & DC$5$-ResNet-$50$ & $ 197 $ & $18748$ & $10$ & $ 44.8 $   \\
         \rowcolor{lightgray} Conditional DETR v2 & DC$5$-ResNet-$50$ & $ 161 $ & $4844$ & $16$ & $ 44.8 $   \\
        \shline
    \end{tabular}
    \vspace{-0.1cm}
\end{table}

\begin{table}[ht]
    \centering
    \setlength{\tabcolsep}{10pt}
        \scriptsize
            \renewcommand{\arraystretch}{1.3}
    \caption{\small Inference speed and AP performance on COCO 2017 \texttt{val}. $\texttt{TorchScript}$ is not used here. 
    All models are trained with $50$ epochs except DETR that is trained with 500 epochs. Our model achieves the best speed/accuracy trade-off.}
    \vspace{-0.1cm}
    \label{tab:abla-speed}
    \begin{tabular}{ l c  c}
    \shline
         Method & Inference Speed (FPS) & AP \\
         \shline
         DETR & $10$ & $ 43.3 $ \\
         SMCA & $10$ & $ 43.7 $ \\
         Conditional DETR & $10$ & $ 43.8 $ \\
         Anchor DETR & $16$ & $ 44.2 $ \\
         \rowcolor{lightgray} Conditional DETR v2 & $16$ & $ 44.8 $   \\
        \shline
    \end{tabular}
    \vspace{-0.1cm}
\end{table}

\renewcommand{\baselinestretch}{1}
\selectfont

\vspace{.1cm}
\noindent\textbf{Analysis of the computational cost and the speed.}
We compare the computational cost and speed of the proposed method to Conditional DETR in Table~\ref{tab:abla-compute-cost}. For the ResNet-$50$ backbone, the proposed Horizontal-Vertical Attention reduces the memory cost significantly, but only 1G flops are saved. This is because the overall computational cost of the transformer head is relatively small for the $32\times$ feature. 
For the DC$5$-ResNet-$50$ backbone, the global self-attention consumes huge resources, $e.g., 18748$ M memory. The speed is much slower due to the high computational complexity. With the help of the proposed Horizontal-Vertical Attention, the memory cost is reduced by $74$\% and the speed is increased by $60$\%. We also compare the efficiency to other DETR variants in Table~\ref{tab:abla-speed}, where we could find that our method achieves the best speed/accuracy trade-off.

\section{Conclusion}
We reformulate the object query in DETR to the format of the box query, which is a composition of the embedding of the reference point and the transformation of the box with respect to the reference point. We further learn the box query from the image content instead of setting them as the model parameters, which improves the detection performance. To save the memory cost of the transformer encoder, we learn from axial self-attention and propose to perform attention in horizontal and vertical directions in parallel. Experiments verify the efficiency of the proposed method.

\clearpage

\appendix

\section{Visualization of the Reference Points}
We select reference points from the image content, as illustrated in Section $3.2$ in the main paper. We visualize the reference points in Figure~\ref{fig:reference-points} and find that the reference points mainly exist in the object regions. For example, in the last image of the second row, almost all the reference points are distributed on the three cows. 

\begin{center}
\centering
\includegraphics[width=\textwidth]{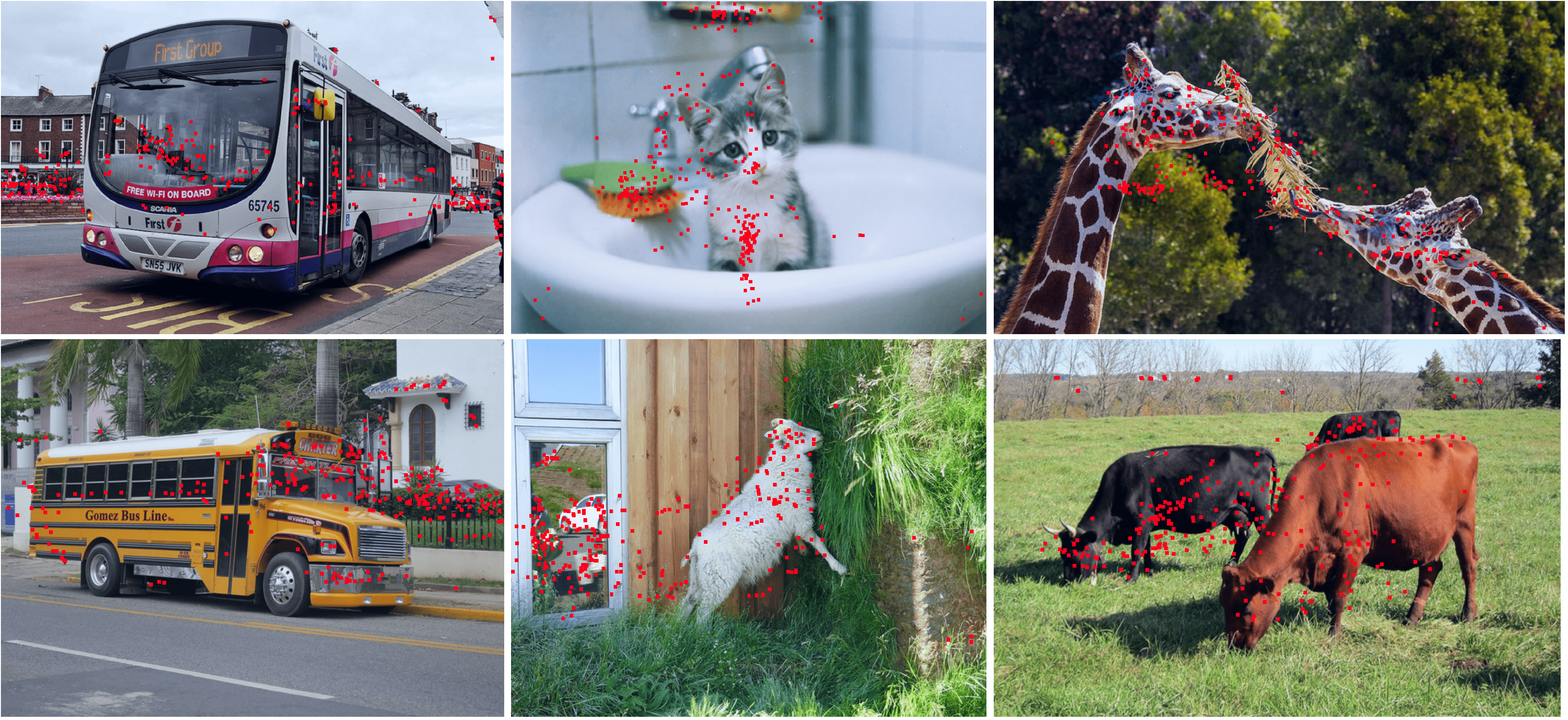}
   \captionof{figure}{
  Visualization of the reference points. We use $300$ reference points in the model, which are marked with red in the images. Best viewed in color.}
\label{fig:reference-points}
\end{center}

\section{Approach Details}

\vspace{0.1cm}
\noindent \textbf{Decoder cross attention weight.} For the internal decoder layers, we calculate the decoder cross attention weight in the same way as Conditional DETR~\cite{meng2021conditional}:
\begin{align}
\label{eqn:internal-conditional-attention}
&\mathbf{c}_q^\top \mathbf{c}_k + \mathbf{p}_q^\top \mathbf{p}_k \nonumber \\ 
=~& \mathbf{c}_q^\top \mathbf{c}_k + ({\uplambda}_q \odot {\mathbf{p}}_s)^\top \mathbf{p}_k,
\end{align}
where $\mathbf{c}_q, \mathbf{c}_k, \mathbf{p}_q, \mathbf{p}_k$ are the content query, content key, spatial query and spatial key, respectively. The content query consists of the embedding $(\mathbf{p}_s)$ of the reference point, and the image-dependent transformation $({\uplambda}_q)$. 

For the first decoder layer, 
the cross attention weight of Conditional DETR~\cite{meng2021conditional} is calculated by:
\begin{align}
    (\mathbf{c}_q + \mathbf{o}_q)^\top (\mathbf{c}_k + \mathbf{p}_k) + \mathbf{o}_q^\top \mathbf{p}_k,
\end{align}

where $\mathbf{o}_q$ is the object query. This is different from Eq. (\ref{eqn:internal-conditional-attention}) since there is no image-dependent information as input in the first decoder layer of Conditional DETR~\cite{meng2021conditional}.

By contrast, we adopt the same formulation for the first decoder layer as Eq. (\ref{eqn:internal-conditional-attention}).
The reference point is initialized from the image content and is likely to lie in the object regions. The sinusoidal embedding is applied to the reference point to obtain $\mathbf{p}_s$. The transformation $({\uplambda}_q)$ that contains the scale information of the object is predicted from the corresponding encoder embedding of the reference point.

%
%
\bibliographystyle{splncs04}
\bibliography{egbib}

\begin{thebibliography}{10}
\providecommand{\url}[1]{\texttt{#1}}
\providecommand{\urlprefix}{URL }
\providecommand{\doi}[1]{https://doi.org/#1}

\bibitem{BPA20}
Beltagy, I., Peters, M.E., Cohan, A.: Longformer: The long-document
  transformer. CoRR  \textbf{abs/2004.05150} (2020)

\bibitem{BWL20}
Bochkovskiy, A., Wang, C., Liao, H.M.: Yolov4: Optimal speed and accuracy of
  object detection. CoRR  \textbf{abs/2004.10934} (2020)

\bibitem{CaiV18}
Cai, Z., Vasconcelos, N.: Cascade {R-CNN:} delving into high quality object
  detection. In: CVPR (2018)

\bibitem{CarionMSUKZ20}
Carion, N., Massa, F., Synnaeve, G., Usunier, N., Kirillov, A., Zagoruyko, S.:
  End-to-end object detection with transformers. In: ECCV (2020)

\bibitem{chen2022context}
Chen, X., Ding, M., Wang, X., Xin, Y., Mo, S., Wang, Y., Han, S., Luo, P.,
  Zeng, G., Wang, J.: Context autoencoder for self-supervised representation
  learning. arXiv preprint arXiv:2202.03026  (2022)

\bibitem{chen20203d}
Chen, X., Lin, K.Y., Qian, C., Zeng, G., Li, H.: 3d sketch-aware semantic scene
  completion via semi-supervised structure prior. In: CVPR. pp. 4193--4202
  (2020)

\bibitem{chen2020bi}
Chen, X., Lin, K.Y., Wang, J., Wu, W., Qian, C., Li, H., Zeng, G.:
  Bi-directional cross-modality feature propagation with
  separation-and-aggregation gate for rgb-d semantic segmentation. In: ECCV.
  pp. 561--577. Springer (2020)

\bibitem{chen2021semi}
Chen, X., Yuan, Y., Zeng, G., Wang, J.: Semi-supervised semantic segmentation
  with cross pseudo supervision. In: CVPR. pp. 2613--2622 (2021)

\bibitem{CGRS19}
Child, R., Gray, S., Radford, A., Sutskever, I.: Generating long sequences with
  sparse transformers. CoRR  \textbf{abs/1904.10509} (2019),
  \url{http://arxiv.org/abs/1904.10509}

\bibitem{Krzysztof20}
Choromanski, K., Likhosherstov, V., Dohan, D., Song, X., Gane, A.,
  Sarl{\'{o}}s, T., Hawkins, P., Davis, J., Mohiuddin, A., Kaiser, L.,
  Belanger, D., Colwell, L., Weller, A.: Rethinking attention with performers.
  CoRR  \textbf{abs/2009.14794} (2020), \url{https://arxiv.org/abs/2009.14794}

\bibitem{DaiCLC20}
Dai, Z., Cai, B., Lin, Y., Chen, J.: {UP-DETR:} unsupervised pre-training for
  object detection with transformers. CoRR  \textbf{abs/2011.09094} (2020),
  \url{https://arxiv.org/abs/2011.09094}

\bibitem{DaiYYCLS19}
Dai, Z., Yang, Z., Yang, Y., Carbonell, J.G., Le, Q.V., Salakhutdinov, R.:
  Transformer-xl: Attentive language models beyond a fixed-length context. In:
  ACL (2019)

\bibitem{GaoZWDL21}
Gao, P., Zheng, M., Wang, X., Dai, J., Li, H.: Fast convergence of {DETR} with
  spatially modulated co-attention. CoRR  \textbf{abs/2101.07448} (2021),
  \url{https://arxiv.org/abs/2101.07448}

\bibitem{GlorotB10}
Glorot, X., Bengio, Y.: Understanding the difficulty of training deep
  feedforward neural networks. In: AISTATS (2010)

\bibitem{HeZRS16}
He, K., Zhang, X., Ren, S., Sun, J.: Deep residual learning for image
  recognition. In: CVPR (2016)

\bibitem{HoKWS19}
Ho, J., Kalchbrenner, N., Weissenborn, D., Salimans, T.: Axial attention in
  multidimensional transformers. CoRR  \textbf{abs/1912.12180} (2019),
  \url{http://arxiv.org/abs/1912.12180}

\bibitem{huang2019ccnet}
Huang, Z., Wang, X., Huang, L., Huang, C., Wei, Y., Liu, W.: Ccnet: Criss-cross
  attention for semantic segmentation. In: Proceedings of the IEEE/CVF
  International Conference on Computer Vision. pp. 603--612 (2019)

\bibitem{KKL20}
Kitaev, N., Kaiser, L., Levskaya, A.: Reformer: The efficient transformer. In:
  ICLR (2020), \url{https://openreview.net/forum?id=rkgNKkHtvB}

\bibitem{law2018cornernet}
Law, H., Deng, J.: Cornernet: Detecting objects as paired keypoints. In:
  Proceedings of the European conference on computer vision (ECCV). pp.
  734--750 (2018)

\bibitem{LinGGHD20}
Lin, T., Goyal, P., Girshick, R.B., He, K., Doll{\'{a}}r, P.: Focal loss for
  dense object detection. TPAMI  (2020)

\bibitem{LinMBHPRDZ14}
Lin, T., Maire, M., Belongie, S.J., Hays, J., Perona, P., Ramanan, D.,
  Doll{\'{a}}r, P., Zitnick, C.L.: Microsoft {COCO:} common objects in context.
  In: ECCV (2014)

\bibitem{LiuAESRFB16}
Liu, W., Anguelov, D., Erhan, D., Szegedy, C., Reed, S.E., Fu, C., Berg, A.C.:
  {SSD:} single shot multibox detector. In: ECCV (2016)

\bibitem{liu2021swin}
Liu, Z., Lin, Y., Cao, Y., Hu, H., Wei, Y., Zhang, Z., Lin, S., Guo, B.: Swin
  transformer: Hierarchical vision transformer using shifted windows. arXiv
  preprint arXiv:2103.14030  (2021)

\bibitem{LoshchilovH17}
Loshchilov, I., Hutter, F.: Fixing weight decay regularization in adam. In:
  ICLR (2017)

\bibitem{meng2021conditional}
Meng, D., Chen, X., Fan, Z., Zeng, G., Li, H., Yuan, Y., Sun, L., Wang, J.:
  Conditional detr for fast training convergence. In: ICCV. pp. 3651--3660
  (2021)

\bibitem{PPYSK21}
Peng, H., Pappas, N., Yogatama, D., Schwartz, R., Smith, N.A., Kong, L.: Random
  feature attention. CoRR  \textbf{abs/2103.02143} (2021),
  \url{https://arxiv.org/abs/2103.02143}

\bibitem{ramachandran2019stand}
Ramachandran, P., Parmar, N., Vaswani, A., Bello, I., Levskaya, A., Shlens, J.:
  Stand-alone self-attention in vision models. arXiv preprint arXiv:1906.05909
  (2019)

\bibitem{RedmonF17}
Redmon, J., Farhadi, A.: {YOLO9000:} better, faster, stronger. In: CVPR (2017)

\bibitem{JA18}
Redmon, J., Farhadi, A.: Yolov3: An incremental improvement. CoRR
  \textbf{abs/1804.02767} (2018)

\bibitem{ren2015faster}
Ren, S., He, K., Girshick, R., Sun, J.: Faster r-cnn: Towards real-time object
  detection with region proposal networks. Advances in neural information
  processing systems  \textbf{28},  91--99 (2015)

\bibitem{RenHG017}
Ren, S., He, K., Girshick, R.B., Sun, J.: Faster {R-CNN:} towards real-time
  object detection with region proposal networks. TPAMI  (2017)

\bibitem{RezatofighiTGS019}
Rezatofighi, H., Tsoi, N., Gwak, J., Sadeghian, A., Reid, I.D., Savarese, S.:
  Generalized intersection over union: {A} metric and a loss for bounding box
  regression. In: CVPR (2019)

\bibitem{RSVG21}
Roy, A., Saffar, M., Vaswani, A., Grangier, D.: Efficient content-based sparse
  attention with routing transformers. ACL  (2021),
  \url{https://transacl.org/ojs/index.php/tacl/article/view/2405}

\bibitem{SunCYK20}
Sun, Z., Cao, S., Yang, Y., Kitani, K.: Rethinking transformer-based set
  prediction for object detection. CoRR  \textbf{abs/2011.10881} (2020),
  \url{https://arxiv.org/abs/2011.10881}

\bibitem{tang2022not}
Tang, J., Chen, X., Wang, J., Zeng, G.: Not all voxels are equal: Semantic
  scene completion from the point-voxel perspective. In: AAAI. vol.~36, pp.
  2352--2360 (2022)

\bibitem{tang2022point}
Tang, J., Chen, X., Wang, J., Zeng, G.: Point scene understanding via
  disentangled instance mesh reconstruction. In: ECCV (2022)

\bibitem{TianSCH19}
Tian, Z., Shen, C., Chen, H., He, T.: {FCOS:} fully convolutional one-stage
  object detection. In: ICCV (2019)

\bibitem{vaswani2021scaling}
Vaswani, A., Ramachandran, P., Srinivas, A., Parmar, N., Hechtman, B., Shlens,
  J.: Scaling local self-attention for parameter efficient visual backbones.
  In: CVPR. pp. 12894--12904 (2021)

\bibitem{VaswaniSPUJGKP17}
Vaswani, A., Shazeer, N., Parmar, N., Uszkoreit, J., Jones, L., Gomez, A.N.,
  Kaiser, L., Polosukhin, I.: Attention is all you need. In: NeurIPS (2017)

\bibitem{WLKFM20}
Wang, S., Li, B.Z., Khabsa, M., Fang, H., Ma, H.: Linformer: Self-attention
  with linear complexity. CoRR  \textbf{abs/2006.04768} (2020),
  \url{https://arxiv.org/abs/2006.04768}

\bibitem{wang2021pnp}
Wang, T., Yuan, L., Chen, Y., Feng, J., Yan, S.: Pnp-detr: Towards efficient
  visual analysis with transformers. In: ICCV. pp. 4661--4670 (2021)

\bibitem{wang2021anchor}
Wang, Y., Zhang, X., Yang, T., Sun, J.: Anchor detr: Query design for
  transformer-based detector. arXiv preprint arXiv:2109.07107  (2021)

\bibitem{XiongZCTFLS21}
Xiong, Y., Zeng, Z., Chakraborty, R., Tan, M., Fung, G., Li, Y., Singh, V.:
  Nystr{\"{o}}mformer: {A} nystr{\"{o}}m-based algorithm for approximating
  self-attention. CoRR  \textbf{abs/2102.03902} (2021),
  \url{https://arxiv.org/abs/2102.03902}

\bibitem{yuan2021hrformer}
Yuan, Y., Fu, R., Huang, L., Lin, W., Zhang, C., Chen, X., Wang, J.: Hrformer:
  High-resolution transformer for dense prediction. arXiv preprint
  arXiv:2110.09408  (2021)

\bibitem{ZGDAAOPRWYA20}
Zaheer, M., Guruganesh, G., Dubey, K.A., Ainslie, J., Alberti, C.,
  Onta{\~{n}}{\'{o}}n, S., Pham, P., Ravula, A., Wang, Q., Yang, L., Ahmed, A.:
  Big bird: Transformers for longer sequences. In: NeurIPS (2020),
  \url{https://proceedings.neurips.cc/paper/2020/hash/c8512d142a2d849725f31a9a7a361ab9-Abstract.html}

\bibitem{ZWP19}
Zhou, X., Wang, D., Kr{\"{a}}henb{\"{u}}hl, P.: Objects as points. CoRR
  \textbf{abs/1904.07850} (2019), \url{http://arxiv.org/abs/1904.07850}

\bibitem{zhou2019bottom}
Zhou, X., Zhuo, J., Krahenbuhl, P.: Bottom-up object detection by grouping
  extreme and center points. In: Proceedings of the IEEE/CVF conference on
  computer vision and pattern recognition. pp. 850--859 (2019)

\bibitem{ZhuSLLWD20}
Zhu, X., Su, W., Lu, L., Li, B., Wang, X., Dai, J.: Deformable {DETR:}
  deformable transformers for end-to-end object detection. CoRR
  \textbf{abs/2010.04159} (2020), \url{https://arxiv.org/abs/2010.04159}

\end{thebibliography}
\end{document}